\documentclass[10pt,twocolumn,letterpaper]{article}

\usepackage[pagenumbers]{cvpr} 

\usepackage[dvipsnames]{xcolor}

\definecolor{cvprblue}{rgb}{0.21,0.49,0.74}
\usepackage[pagebackref=false,breaklinks,colorlinks,citecolor=cvprblue]{hyperref}

\usepackage{multirow}
\usepackage{multicol}
\usepackage{xcolor, colortbl}
\usepackage{algorithm}
\usepackage{amsfonts} 
\usepackage{listings}
\definecolor{citecolor}{HTML}{0071bc}
\definecolor{codeblue}{rgb}{0.25,0.5,0.5}
\definecolor{myblue}{rgb}{0.88,0.98,0.98}
\definecolor{mygreen}{rgb}{0, 0.5, 0.1}
\definecolor{myred}{rgb}{0.84, 0.04, 0.33}
\definecolor{mygray}{gray}{0.92}
\usepackage{pifont}
\newcommand{\cmark}{\ding{51}}%
\newcommand{\xmark}{\ding{55}}%
\newcolumntype{P}[1]{>{\centering\arraybackslash}p{#1}}
\usepackage{etoolbox}
\makeatletter
\AfterEndEnvironment{algorithm}{\let\@algcomment\relax}
\AtEndEnvironment{algorithm}{\kern2pt\hrule\relax\vskip3pt\@algcomment}
\let\@algcomment\relax
\newcommand\algcomment[1]{\def\@algcomment{\footnotesize#1}}
\makeatother

\def\paperID{1234} 
\def\confName{CVPR}
\def\confYear{2024}

\title{LLaMA-VID: An Image is Worth 2 Tokens in Large Language Models}

\author{
Yanwei Li$^{1}$\thanks{ equal contribution}~\quad Chengyao Wang$^{1*}$~\quad Jiaya Jia$^{1,2}$ \\[0.2cm]
CUHK$^{1}$\quad SmartMore$^{2}$
}

\begin{document}
\maketitle

\begin{abstract}
In this work, we present a novel method to tackle the token generation challenge in Vision Language Models (VLMs) for video and image understanding, called LLaMA-VID. 
Current VLMs, while proficient in tasks like image captioning and visual question answering, face computational burdens when processing long videos due to the excessive visual tokens. 
LLaMA-VID addresses this issue by representing each frame with two distinct tokens, namely context token and content token. 
The context token encodes the overall image context based on user input, whereas the content token encapsulates visual cues in each frame. 
This dual-token strategy significantly reduces the overload of long videos while preserving critical information. 
Generally, LLaMA-VID empowers existing frameworks to support hour-long videos and pushes their upper limit with an extra context token. 
It is proved to surpass previous methods on most of video- or image-based benchmarks. 
Code is available at~\href{https://github.com/dvlab-research/LLaMA-VID}{https://github.com/dvlab-research/LLaMA-VID}.

\end{abstract}

\section{Introduction}
Large Language Models (LLMs)~\cite{ChatGPT,zhang2022opt,llama}, through their capacity to generate contextually accurate responses, have significantly advanced the field of AI.
Drawing from the strengths of LLMs, Vision Language Models (VLMs)~\cite{instructblip,llava,GPT4} have been developed to extend these capabilities to visual data, demonstrating their adeptness in tasks like image captioning and visual question answering. 
However, a substantial challenge emerges in the context of long video, where an excessive number of tokens are required to represent consecutive frames. 
The computational demands escalate with the video length, thereby constraining the practical application of VLMs for extensive videos.

Recently, several approaches have been proposed to handle videos, moving beyond image-only VLMs.
These methods aim to alleviate the token issue by utilizing representative queries~\cite{videochat,videollama} or applying temporal compression~\cite{videochatgpt,luo2023valley}.
Despite these efforts, the challenge of long videos remains unresolved. 
The primary obstacle stems from the excessive number of tokens required for each video frame. 
For instance, models like BLIP~\cite{blip2,instructblip} and LLaVA~\cite{llava} require 32 and over 256 tokens respectively for a single image. 
A video containing 10K frames would thus necessitate over 320K tokens, exceeding the capacity of current VLMs. 
Furthermore, simple temporal compression can significantly damage the representation over long-term intervals. 
This drawback hampers their performance, thereby underscoring the need for a robust solution.

\begin{figure}[t!]
\centering
\includegraphics[width=0.89\linewidth]{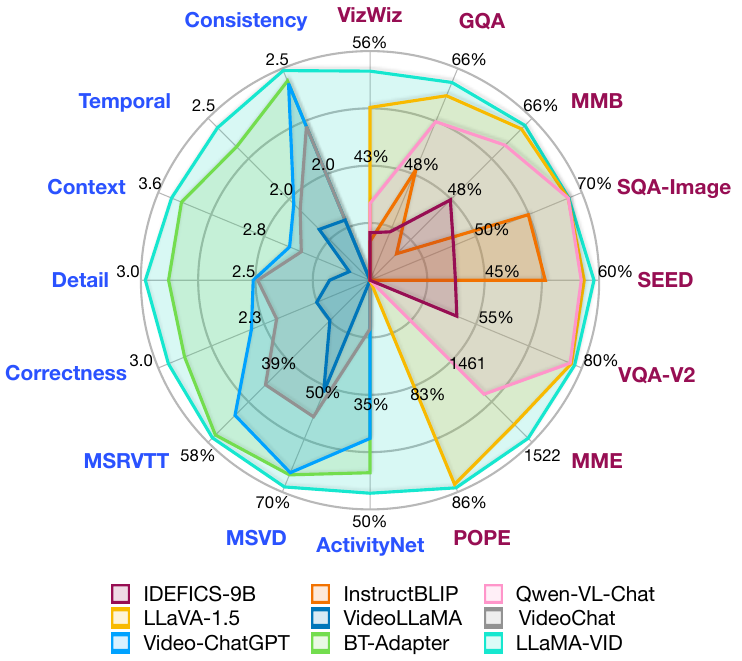} 
\caption{
The proposed LLaMA-VID achieves leading performance on most of benchmarks with 7B LLMs. The video-based and image-based benchmarks are noted in \textcolor{blue}{blue} and \textcolor{purple}{purple} color, respectively.
Please refer to Tables~\ref{tab:main_video},~\ref{tab:main_video_gen}, and~\ref{tab:main_img} for more details.
}
\label{fig:intro}
\end{figure}

In this work, we present LLaMA-VID, a novel approach to effectively manage the token generation issue in long videos. 
Our core idea is to represent each video frame with two distinct tokens: {\em context token} and {\em content token}. 
The context token is designed to encode the overall context of the image based on user input, which efficiently condenses the broader picture into {\em a single token}. 
Simultaneously, the content token captures finer aspects of each frame. 
According to computational constraints, the length of content token can be extended to include more details, {\em e.g.}, 1 token/frame for video input and beyond 256 token/frame for single image. 
In this way, the overload of long videos can be significantly reduced without sacrificing critical information.

In particular, our method employs a dual-token generation strategy that is both efficient and effective. 
For each frame, we first extract image features using a pre-trained vision transformer~\cite{vit}, akin to other VLMs~\cite{instructblip,llava}. 
The key question is how to generate the context-related token according to user instructions.
We provide the solution by leveraging the cross-modality design~\cite{devlin2018bert,blip2} for instruction-guided queries, which carry the interactive intention from users.
For {\em context token}, these queries interact with previously generated image features in the designed attention module, termed as context attention.
To generate {\em content token}, the image features are average pooled to formulate tokens that adapt to different settings.
For instance, global pooling is adopted to maintain efficiency for video input while details are preserved with more tokens for single image input. 
The context and content tokens are subsequently projected to the space of LLMs with simple linear layers for final prediction.
Furthermore, to better support hour-long videos in VLMs, we construct an instruction-based dataset that contains 9K movie-level conversations for plot reasoning and detail understanding.

Generally, LLaMA-VID can be distinguished from two aspects.
On one hand, with the dual-token paradigm, each frame can be efficiently encoded with only two tokens, which empowers existing LLMs to support long videos.
On the other hand, the context token aggregates the most informative feature of each image, which further extends the upper limit of VLMs with an extra token. 

The overall framework, dubbed LLaMA-VID, can be easily instantiated with various decoders and LLMs, as elaborated in Section~\ref{sec:method}.
Extensive empirical studies are conducted in Section~\ref{sec:experiment} to reveal the effectiveness of each component.
Remarkably, our model can complete training within 2 days on a single machine with 8$\times$A100 GPUs, and it outperforms previous leading methods on most of video- and image-based benchmarks, as shown in Figure~\ref{fig:intro}.

\section{Related Work}~\label{sec:related}
In this section, we first review large language models and delve into recent advances in vision language models.
\subsection{Large Language Models}
The field of Natural Language Processing (NLP) has witnessed tremendous advancements with the evolution of LLMs.
Transformer~\cite{vaswani2017attention} marked a pivotal milestone, with subsequent language models~\cite{devlin2018bert,liu2019roberta,zhang2022opt} demonstrating remarkable capabilities.
GPT~\cite{brown2020language} revolutionized this field by utilizing generative pre-trained transformers for auto-regressive prediction, which is proved to be a potent language modeling paradigm. 
Recent groundbreaking works, such as ChatGPT~\cite{ChatGPT}, GPT-4~\cite{GPT4}, and LLaMA~\cite{llama}, have pushed the boundaries even further. 
Trained on vast amounts of text data, these models exhibit exceptional capabilities in complex linguistic tasks.
To leverage the potential of pre-trained LLMs, instruction tuning~\cite{wei2021finetuned,ouyang2022training} is a crucial component for high-quality output.
This strategy is widely adopted in open-source models like Alpaca~\cite{alpaca} and Vicuna~\cite{vicuna}, which improve over LLaMA~\cite{llama} using specially designed instruction pairs. 
There are also researches~\cite{visualchatgpt,gpt4tools} that utilize the reasoning ability of LLMs and invoke pre-defined tools for visual applications.
Different from them, we collect multi-modality instruction data that contains text, images, and videos in this work, which is employed to empower LLMs for long video processing.

\subsection{Vision Language Models}
The advancements in computer vision and NLP have led to the emergence of vision-language models (VLMs) that integrate vision models with language models for cross-modality understanding~\cite{cococap,msrvtt} and reasoning~\cite{vqav2,scienceqa,lai2023lisa}.
Pioneering large-scale VLMs like CLIP~\cite{CLIP} and ALIGN~\cite{ALIGN} have extended language models to vision-language tasks. 
The recent progress has seen an increasing focus on leveraging the power of LLMs.
Notably, Flamingo~\cite{flamingo} and BLIP-2~\cite{blip2} utilize web-scale image-text pairs for cross-modality alignment, thereby enhancing learning performance.
To further exploit the potential of such pre-trained models, InstructBLIP~\cite{instructblip} and MiniGPT-4~\cite{minigpt4} construct high-quality instruction pairs based on BLIP-2 and achieve superior results. 
Simultaneously, LLaVA~\cite{llava} employs a simple linear projector with a few learnable parameters to align the image and text space of LLaMA.
Given the tailored instruction data, this straightforward approach demonstrates strong capabilities.
To support video understanding in LLMs, Video-LLaMA~\cite{videollama} and VideoChat~\cite{videochat} attempt to utilize BLIP-2 for video embedding extraction, while Video-ChatGPT~\cite{videochatgpt} proposes spatial and temporal pooling for video features.
However, given the substantial number of tokens required for each frame, LLMs encounter significant challenges when processing extensive video sequences. 
It prevents previous work from representing long video sequences that exceed a duration of one hour in LLMs.
To solve the issue, we propose to efficiently encode each frame with only 2 tokens, which supports long video understanding in existing LLMs.

\begin{figure*}[t!]
\centering
\includegraphics[width=0.98\linewidth]{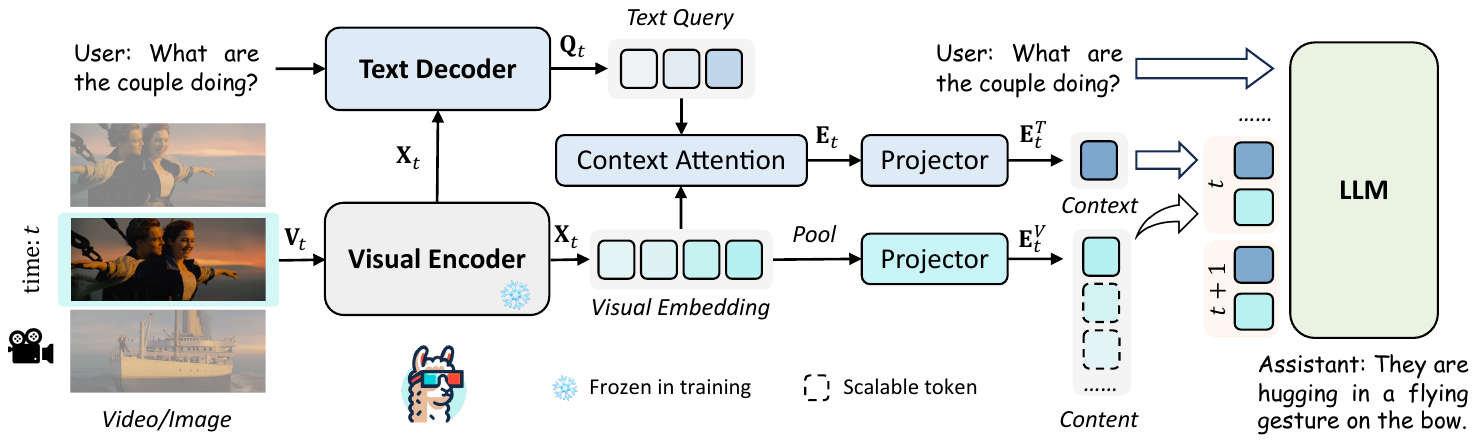} 
\caption{
The framework of LLaMA-VID.
With user directive, LLaMA-VID operates by taking either a {\em single image} or {\em video frames} as input, and generates responses from LLM.
The process initiates with a visual encoder that transforms input frames into the visual embedding.
Then, the text decoder produces text queries based on the user input.
In context attention, the text query aggregates text-related visual cues from the visual embedding.
For efficiency, an option is provided to downsample the visual embedding to various token sizes, or even to a single token.
The text-guided {\em context token} and the visually-enriched {\em content token} are then formulated using a linear projector to represent each frame at time $t$.
Finally, the LLM takes the user directive and all visual tokens as input and gives responses.
}
\label{fig:main}
\end{figure*}

\section{LLaMA-VID}~\label{sec:method}
The framework of LLaMA-VID is conceptually simple: encoder and decoder are adopted to produce visual embedding and text-guided features, respectively; 
context token and content token are transformed with the tailored token generation strategy; 
instruction tuning is designed to unleash the potential of LLMs for image and video.

\subsection{Encoder and Decoder}~\label{sec:sub_encoder}
The proposed LLaMA-VID can be utilized to interact with single image or long videos.
For clarity, we assume the input image is captured from a video sequence, as presented in Figure~\ref{fig:main}.
Given a video frame ${\mathbf V}_t\in\mathbb{R}^{H\times W\times 3}$ at time $t$, a transformer-based visual encoder is first employed to produce the visual embedding  ${\mathbf X}_t\in\mathbb{R}^{N\times C}$. Here, $N=H/p\times W/p$ and $C$ indicate the number of image patches and embedding channels, respectively.
The patch size $p$ is typically set to 14 for ViT-based backbones~\cite{vit,CLIP,evaclip}.
Meanwhile, we take the user instruction as input and generate the text-guided query ${\mathbf Q}_t\in\mathbb{R}^{M\times C}$ with the produced ${\mathbf X}_t$, where $M$ denotes the number of queries.
As depicted in Figure~\ref{fig:main}, this cross-modality interaction predominantly occurs in the text decoder, which can be easily instantiated with BERT~\cite{devlin2018bert} or QFormer~\cite{instructblip}, as compared in Table~\ref{tab:abla_text_vis}.
In this way, the text query ${\mathbf Q}_t$ contains highlighted visual cues that are most related to the user instruction.

\subsection{Token Generation}~\label{sec:sub_token}
With the text query ${\mathbf Q}_t$ and visual embedding ${\mathbf X}_t$, we can easily generate representative tokens for LLMs.
Specifically, context attention is designed to aggregate text-related visual features and condense them to a single context token.
As shown in Figure~\ref{fig:main}, it takes ${\mathbf Q}_t$ and ${\mathbf X}_t$ as input and formulates the context-related embedding ${\mathbf E}_t\in\mathbb{R}^{1\times C}$ as

\begin{algorithm}[t!]
\caption{Pseudo Code for Token Generation.}
\label{algo:code}
\algcomment{\fontsize{7.2pt}{0em}\selectfont \texttt{F}: torch.nn.functional; \texttt{ctxproj}, \texttt{visproj}: predefined linear projectors.
}
\definecolor{codeblue}{rgb}{0.25,0.5,0.5}
\lstset{
  backgroundcolor=\color{white},
  basicstyle=\fontsize{7.2pt}{7.2pt}\ttfamily\selectfont,
  columns=fullflexible,
  breaklines=true,
  captionpos=b,
  commentstyle=\fontsize{7.2pt}{7.2pt}\color{codeblue},
  keywordstyle=\fontsize{7.2pt}{7.2pt},
}
\begin{lstlisting}[language=python]
# B: batch size; C: channel size; n: content shape
# M: query length; N: shape of flatten image pacthes; 
# text_q: text query in shape (B, M, C)
# vis_embed: visual embedding in shape (B, N, C)

# Key part 1: calculate context-related embedding
ctx_embed = text_q @ vis_embed.transpose(-1,-2)
ctx_embed = ctx_embed / (vis_embed.shape[-1]**0.5)
ctx_embed = (ctx_embed.softmax(-1)@vis_embed).mean(1)
ctx_embed = self.ctxproj(ctx_embed[:,None])

# Key part 2: calculate visual embedding
cur_shape = int(vis_embed.shape[1]**0.5)
vis_embed = vis_embed.reshape(B, cur_shape, -1, C)
vis_embed = F.avg_pool2d(vis_embed.permute(0,3,1,2), kernel_size=cur_shape//n, stride=cur_shape//n)
vis_embed = vis_embed.permute(0,2,3,1).flatten(1,2)
vis_embed = self.visproj(vis_embed)

# concat token in shape (B, n+1, C), n in [1,N]
final_token = torch.cat([ctx_embed, vis_embed], dim=1)
\end{lstlisting}
\end{algorithm}

\begin{equation}\label{equ:context_att}
{\mathbf E}_t = {\mathrm {Mean}}({\mathrm {Softmax}}({\mathbf Q}_t \times {\mathbf X}^{\mathsf{T}}_t) \times {\mathbf X}_t),
\end{equation}
where the $\mathrm {Softmax}$ function and $\mathrm {Mean}$ operation are conducted along the $N$ and $M$ dimensions, respectively.
Unlike QFormer~\cite{instructblip} that adopts 32 visual queries as LLMs tokens, we only utilize the text query ${\mathbf Q}_t$ to aggregate the visual features with high-response scores to input instructions.
As a result, the most crucial visual cues related to user input are efficiently preserved in the condensed embedding ${\mathbf E}_t$. 
The effectiveness of this context-related token generation is demonstrated in Table~\ref{tab:abla_token_type} and Figure~\ref{fig:heatmap}.
Subsequently, a linear projector is utilized to transform the embedding ${\mathbf E}_t$ into the context token ${\mathbf E}^{T}_t \in\mathbb{R}^{1\times C}$, which aligns with the language space of LLMs. 
Meanwhile, we employ an adaptive pooling strategy for the visual embedding according to computational constraints to produce the content token ${\mathbf E}^{V}_t \in\mathbb{R}^{n\times C}$, where $n\in[1,N]$.
For instance, we maintain the original resolution of visual embedding ${\mathbf X}_t$ when input single image, while we downsample ${\mathbf X}_t$ to 1 token for long videos. 
This approach significantly reduces the overload of LLMs for each frame, thereby supporting hour-long videos effectively.
Finally, the generated context token ${\mathbf E}^{T}_t$ and the content token ${\mathbf E}^{V}_t$ are concatenated to represent the frame at time $t$. Along with frames at other timestamps, the entire video sequence is translated into the language space in token format, which is then used to generate responses from LLMs. 
The whole process is summarized in Algorithm~\ref{algo:code}.

\begin{figure}[t!]
\centering
\includegraphics[width=\linewidth]{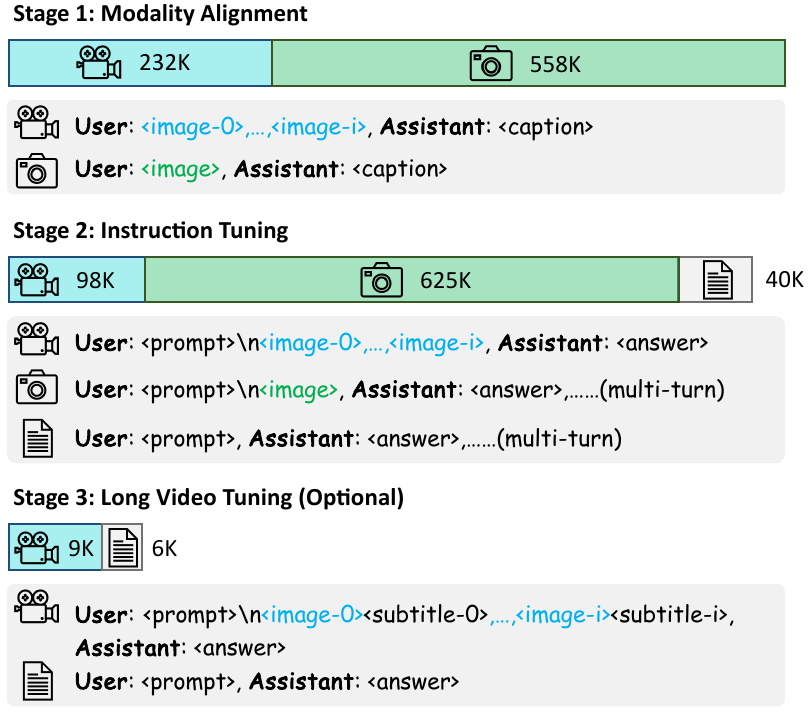} 
\caption{
Multimodal data distribution and instruction format for model training in each stage.
$\texttt{<image>}$ and $\texttt{<image-i>}$ denote the token for single image and the $i$-th video frame, respectively.
}
\label{fig:data}
\end{figure}

\subsection{Training Strategy}~\label{sec:sub_train}
Training strategy, particularly instruction tuning, has proven to be crucial in LLMs~\cite{llama,alpaca,vicuna} and VLMs~\cite{instructblip,llava,llava1.5}.
Considering training efficiency, in this work, we divide the training procedure into three stages, {\em i.e.}, modality alignment, instruction tuning, and long video tuning.

\begin{figure}[t!]
\centering
\includegraphics[width=\linewidth]{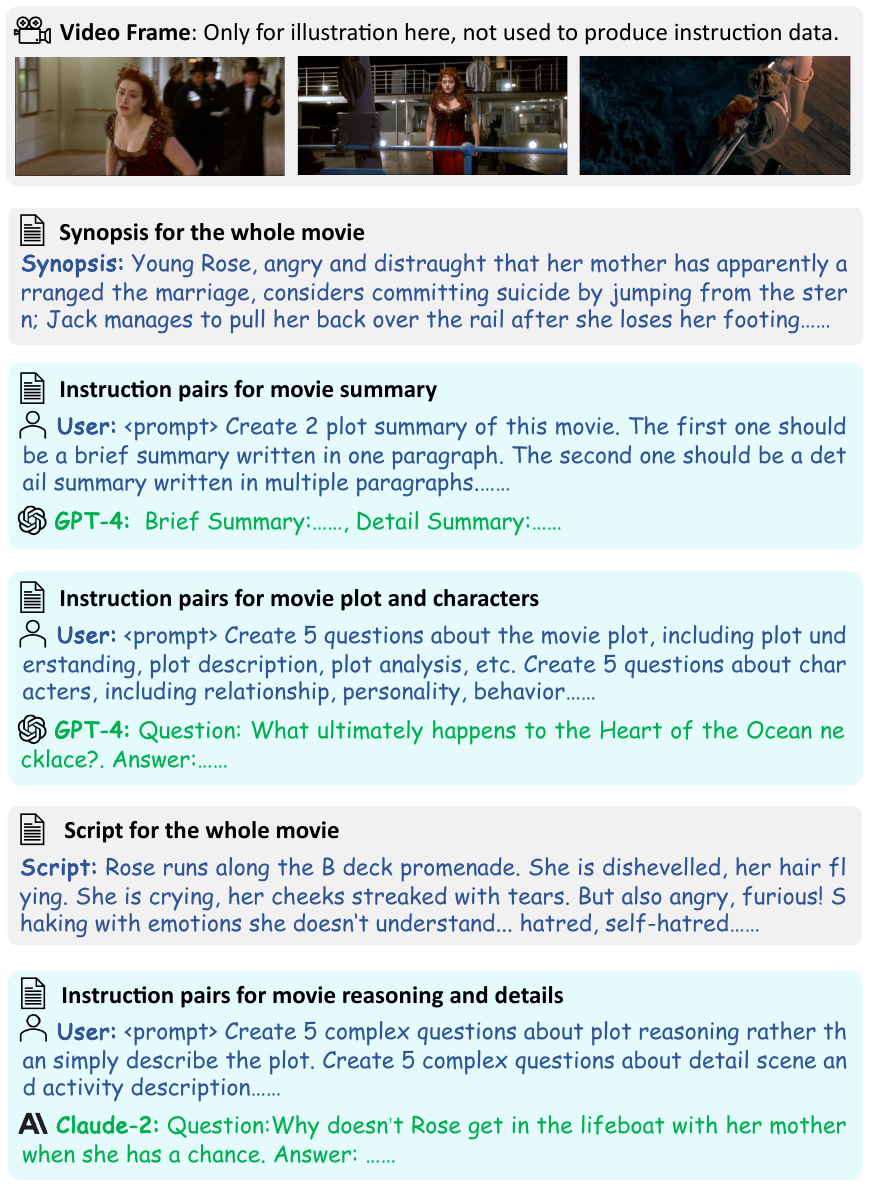} 
\caption{
An example to construct instruction pairs for the movie Titanic.
Given the movie synopsis and script, we utilize the developed LLMs like GPT-4~\cite{GPT4} and Claude-2~\cite{Claude2} to generate movie summaries, plot-related QA pairs, and general reasoning QA pairs.
}
\label{fig:long_movie}
\end{figure}

\vspace{0.5em}
\noindent
\textbf{Modality Alignment.} 
As shown in Figure~\ref{fig:main}, each video frame is projected into the space of LLMs in each forward pass.
Therefore, it is crucial to ensure visual features are well aligned with the language space.
To this end, we construct a compact dataset that contains 790K high-quality image- and video-caption pairs.
As illustrated in Figure~\ref{fig:data}, it mainly consists of 558K image-caption pairs from the LLaVA-filtered CC3M dataset~\cite{cc3m} and 232K video-caption pairs sampled from the WebVid 2.5M dataset~\cite{webvid}.
The instruction format for each modality is presented in Figure~\ref{fig:data}.
In this stage, we primarily optimize the context attention and projectors in Figure~\ref{fig:main}, while freezing the pre-trained modules like the visual encoder and text decoder.

\vspace{0.5em}
\noindent
\textbf{Instruction Tuning.} 
To enhance the multi-modality understanding of LLMs, we build the instruction pairs from~\cite{llava1.5} and~\cite{videochatgpt}.
In particular, the constructed dataset mainly involves content from three sources, {\em i.e.}, 40K text conversations from ShareGPT~\cite{ShareGPT}, 625K single- or multi-turn visual QA pairs from~\cite{llava,vqav2,gqa,okvqa,ocrvqa,aokvqa,textcaps,referitgame,refcoco,vg}, and 98K video QA pairs from~\cite{activitynet}.
For the instruction, we adopt different formats for text, image, and video input, as illustrated in Figure~\ref{fig:data}.
And the input prompt $\texttt{<prompt>}$ and answer $\texttt{<answer>}$ vary with datasets.
Please refer to~\cite{llava1.5} and~\cite{videochatgpt} for more details.
Meanwhile, the image token $\texttt{<image-i>}$ is randomly inserted at the beginning or end of the user input during our training.
In instruction tuning, all the modules are optimized except the frozen visual encoder.

\begin{table*}[t!]

 \centering
\begin{tabular}{llc|ccccccccccc}
  \toprule
  \multirow{2}{*}{Method} & \multirow{2}{*}{LLM} & \multirow{2}{*}{Res.} & \multicolumn{2}{c}{\bf MSVD-QA} & & \multicolumn{2}{c}{\bf MSRVTT-QA} & & \multicolumn{2}{c}{\bf ActivityNet-QA} \\ \cline{4-5} \cline{7-8} \cline{10-11}
  & & & Acc & Score & & Acc & Score & & Acc & Score \\
  \midrule
  FrozenBiLM~\cite{frozenbilm} & DeBERTa-V2 & 224 & 32.2 & -- & & 16.8 & -- & & 24.7 & -- \\
  VideoLLaMA~\cite{videollama} & Vicuna-7B & 224 & 51.6 & 2.5 & & 29.6 & 1.8 & & 12.4 & 1.1 \\
  LLaMA-Adapter~\cite{llamaadapter} & LLaMA-7B & 224 & 54.9 & 3.1 & & 43.8 & 2.7 & & 34.2 & 2.7 \\
  VideoChat~\cite{videochat} & Vicuna-7B & 224 & 56.3 & 2.8 & & 45.0 & 2.5 & & 26.5 & 2.2 \\
  Video-ChatGPT~\cite{videochatgpt} & Vicuna-7B & 224 & 64.9 & \underline{3.3} & & 49.3 & 2.8 & & 35.2 & 2.7 \\
  BT-Adapter~\cite{btadapter} & Vicuna-7B & -- & 67.5 & {\bf 3.7} & & 57.0 & \underline{3.2} & & 45.7 & \underline{3.2} \\
  \midrule
  \rowcolor{mygray}
  {\bf LLaMA-VID} & Vicuna-7B & 224 & \underline{69.7} & {\bf 3.7} & & \underline{57.7} & \underline{3.2} & & \underline{47.4} & {\bf 3.3} \\
  \rowcolor{mygray}
  {\bf LLaMA-VID} & Vicuna-13B & 224 & {\bf 70.0} & {\bf 3.7} & & {\bf 58.9} & {\bf 3.3} & & {\bf 47.5} & {\bf 3.3} \\
  \bottomrule
\end{tabular}
 \caption{Comparison with leading methods on 4 zero-shot video QA datasets. 
 We report results with 2 tokens for each frame.
 For fair comparisons, our model is trained with data of stage 1 and stage 2 without long video tuning in Figure~\ref{fig:data}.
 Res indicates image resolution.
 }
 \label{tab:main_video}
\end{table*}

\vspace{0.5em}
\noindent
\textbf{Long Video Tuning.} 
To further unleash the potential for hour-long videos, we construct 15K long QA pairs, including 9K conversions in movie scenes and 6K data sampled from LongLoRA~\cite{longlora} for token expanding.
Specifically, we utilize more than 400 long movies and corresponding scripts in MovieNet~\cite{movienet} to build the training set.
The key components for instruction generation are visualized in Figure~\ref{fig:long_movie}.
Generally, the generated dataset includes QA pairs from three aspects: video summary, movie plot, and detail reasoning. 
For video summaries, we collect movie synopses to produce brief and detailed summaries for each movie using developed LLMs like GPT-4~\cite{GPT4}. 
It brings about 1K summary-level instruction pairs in total.
For plot-level data, we take the entire movie synopsis as input and leverage GPT-4~\cite{GPT4} to generate plot-related and character-related QA pairs.
These include plot understanding, description, analysis, character relationship, personality, and behavior. 
In particular, we generate 5 plot-related pairs and 5 character-related pairs for each movie, resulting in 4K plot-level QA data.
As for detail-level data, we feed the long movie script into Claude-2~\cite{Claude2} and generate 5 plot-related reasoning pairs and 5 detail-related descriptions for each movie, which brings 4K pairs in total.
With long videos and the generated pairs, we perform instruction tuning by concatenating visual tokens and subtitle tokens for each frame, as depicted in Figure~\ref{fig:data}.
In this way, LLaMA-VID can well support 64K tokens with more than 3-hour video as input.
Please refer to {\em supplementary material} for more details.

\begin{table*}[t!]
 \centering
\begin{tabular}{llc|ccccc}
  \toprule
  Method & LLM & Res. & {\bf Correctness} & {\bf Detail} & {\bf Context} & {\bf Temporal} & {\bf Consistency} \\
  \midrule
  VideoLLaMA~\cite{videollama} & Vicuna-7B & 224 & 1.96 & 2.18 & 2.16 & 1.82 & 1.79 \\
  LLaMA-Adapter~\cite{llamaadapter} & LLaMA-7B & 224 & 2.03 & 2.32 & 2.30 & 1.98 & 2.15 \\
  VideoChat~\cite{videochat} & Vicuna-7B & 224 & 2.23 & 2.50 & 2.53 & 1.94 & 2.24 \\
  Video-ChatGPT~\cite{videochatgpt} & Vicuna-7B & 224 & 2.40 & 2.52 & 2.62 & 1.98 & 2.37 \\
  BT-Adapter~\cite{btadapter} & Vicuna-7B & -- & 2.68 & 2.69 & 3.27 & 2.34 & 2.46 \\
  \midrule
  \rowcolor{mygray}
  {\bf LLaMA-VID} & Vicuna-7B & 224 & \underline{2.96} & \underline{3.00} & \underline{3.53} & \underline{2.46} & \underline{2.51} \\
  \rowcolor{mygray}
  {\bf LLaMA-VID} & Vicuna-13B & 224 & {\bf 3.07} & {\bf 3.05} & {\bf 3.60} & {\bf 2.58} & {\bf 2.63} \\
  \bottomrule
\end{tabular}
 
 \caption{Comparison with leading methods on the video-based generative performance benchmark~\cite{videochatgpt}.
We report results with 2 tokens for each frame.
For fair comparisons, our model is trained with data of stage 1 and stage 2 without long video tuning in Figure~\ref{fig:data}.
Res indicates image resolution.
{\em Correctness}, {\em Detail}, {\em Context}, {\em Temporal}, and {\em Consistency} indicate the evaluation metric of Correctness of Information, Detail Orientation, Contextual Understanding, Temporal Understanding, and Consistency, respectively.
 }
 \label{tab:main_video_gen}
\end{table*}

\begin{table*}[t!]
\centering
\scalebox{1.0}{
\begin{tabular}{l l p{5mm} | p{8mm}P{10mm}P{8mm}P{8mm}P{8mm}P{8mm}P{8mm}p{8mm} }
\toprule
Method & LLM & Res. & {\bf GQA} & {\bf MMB} & {\bf MME} & {\bf POPE} & {\bf SEED} & {\bf SQA}$^\text{I}$ & {\bf VizWiz} & {\bf VQA}$^\text{v2}$ \\
\midrule
InstructBLIP~\cite{instructblip} & Vicuna-7B & 224 & 49.2 & 36.0 & -- & -- & 53.4 & 60.5 & 34.5 & -- \\
IDEFICS-9B~\cite{IDEFICS} & LLaMA-7B & 224 & 38.4 & 48.2 & -- & -- & -- & -- & 35.5 & 50.9 \\
Qwen-VL$^\dagger$~\cite{bai2023qwen} & Qwen-7B & 448 & 59.3* & 38.2 & -- & -- & 56.3 & 67.1 & 35.2 & 78.8* \\
Qwen-VL-Chat$^\dagger$~\cite{bai2023qwen} & Qwen-7B & 448 & 57.5* &  60.6 & 1487.5 & -- & 58.2 & 68.2 & 38.9 & 78.2* \\
LLaVA-1.5~\cite{llava1.5} & Vicuna-7B & 336 & \underline{62.0}* & \underline{64.3} & \underline{1510.7} & \underline{85.9} &  \underline{58.6} & \underline{66.8} & \underline{50.0} & \underline{78.5}* \\
\midrule
\rowcolor{mygray}
{\bf LLaMA-VID} & Vicuna-7B & 336 & {\bf 64.3}* & {\bf 65.1} & {\bf 1521.4} & {\bf 86.0} & {\bf 59.9} & {\bf 68.3} & {\bf 54.2} & {\bf 79.3}* \\
\midrule
BLIP-2~\cite{blip2} & Vicuna-13B & 224 & 41.0 & -- & 1293.8 & 85.3 & 46.4 & 61.0 & 19.6 & 41.0 \\
InstructBLIP~\cite{instructblip} & Vicuna-13B & 224 & 49.5 & -- & 1212.8 & 78.9 & -- & 63.1 & 33.4 & -- \\
Shikra~\cite{chen2023shikra} & Vicuna-13B & 224 & -- & 58.8 & -- & -- & -- & -- & -- & \underline{77.4}* \\
IDEFICS-80B~\cite{IDEFICS} & LLaMA-65B & 224 & 45.2 & 54.5 & -- & -- & -- & -- & 36.0 & 60.0 \\
LLaVA-1.5~\cite{llava1.5} & Vicuna-13B & 336 & \underline{63.3}* & {\bf 67.7} & \underline{1531.3} & \underline{85.9} & \underline{61.6} & {\bf 71.6} & \underline{53.6} & {\bf 80.0}* \\
\midrule
\rowcolor{mygray}
{\bf LLaMA-VID} & Vicuna-13B & 336 & {\bf 65.0}* & \underline{66.6} & {\bf 1542.3} & {\bf 86.0} & {\bf 62.3} & \underline{70.0} & {\bf 54.3} & {\bf 80.0}* \\
\bottomrule
\end{tabular}
}
\caption{Comparison with leading methods on 8 benchmarks. 
Here, we use the same training and instruction finetuning data as that in LLaVA-1.5.
We report results with 1 context token and $n$ content tokens, where $n$ is kept the same with that in LLaVA-1.5, {\em i.e.}, $n=(336/14)^2=576$.
For fair comparisons, our model is trained without video data of stage 1 and stage 2 in Figure~\ref{fig:data}.
Res indicates input image resolution.
$^*$ and $^\dagger$ denote the {\em train} subset is included for training and the data is not publicly available, respectively.}
\label{tab:main_img}
\end{table*}

\section{Experiments}~\label{sec:experiment}
In this section, we provide the experimental setup and comparisons with leading methods on several benchmarks.
More details are attached in {\em supplementary material}.

\subsection{Experimental Setup}
\noindent
\textbf{Implementation Details.}
In this work, we instantiate the model with the pre-trained EVA-G~\cite{evaclip} for visual encoder and QFormer~\cite{instructblip} for text decoder by default.
During training, we keep the visual encoder fixed in all stages and freeze the text decoder, as well as the LLM, in the modality alignment stage, except for the BERT module in Table~\ref{tab:abla_text_vis} that is not pre-trained.
Following the strategy in~\cite{llava1.5}, we optimize trainable parameters with the designed data and instructions in Figure~\ref{fig:data}, running for 1 epoch in each stage.
For video input, we extract frames at a speed of 1 FPS.
All models are trained using 8$\times$NVIDIA A100 GPUs.
Additional hyperparameters are provided in the {\em supplementary material}.

\vspace{1.0em}
\noindent
\textbf{Datasets.}
In this study, we construct the training set mainly from~\cite{llava1.5,webvid,videochatgpt,movienet}, as illustrated in Section~\ref{sec:sub_train}.
Moreover, we report results on several video- and image-based benchmarks.
In particular, for video input, we evaluate the zero-shot performance on the open-ended QA benchmarks like MSVD~\cite{msvd}, MSRVTT~\cite{msrvtt}, ActivityNet~\cite{activitynet}, and the newly-proposed generative performance benchmark~\cite{videochatgpt}.
As for image-based evaluation, we conduct experiments on several widely-adopted benchmarks, including GQA~\cite{gqa}, MMB (MMBench)~\cite{mmbench}, MME~\cite{mme}, POPE~\cite{pope}, SEED~\cite{seed}, SQA$^\text{I}$ (Image-based setting in ScienceQA)~\cite{scienceqa}, VQA$^\text{T}$ (TextVQA)~\cite{textvqa}, VizWiz~\cite{vizwiz}, and VQA$^\text{v2}$ (VQA V2)~\cite{vqav2}.

\begin{figure*}[t!]
\centering
\includegraphics[width=\linewidth]{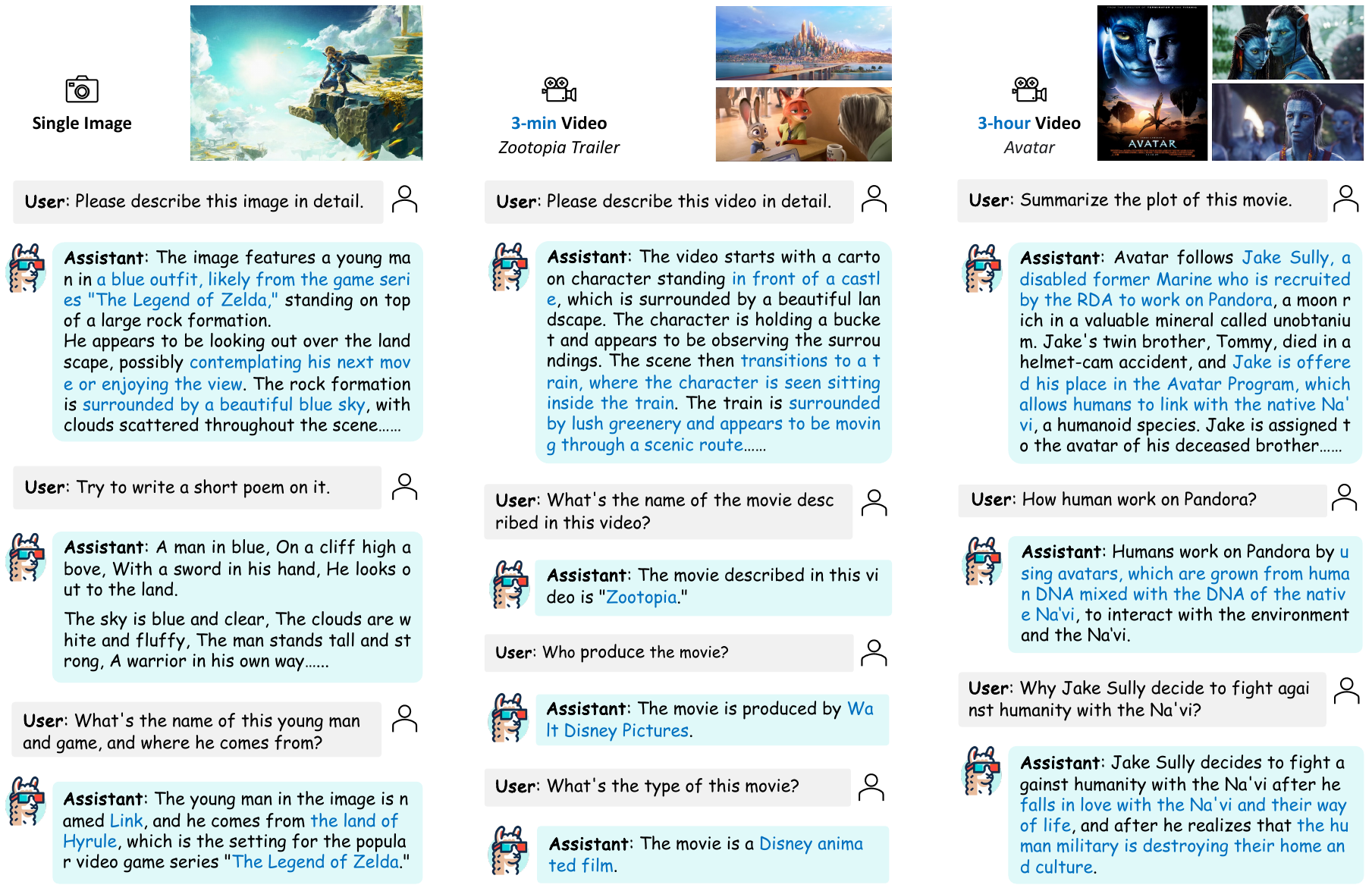} 
\caption{
Examples of LLaMA-VID with single image, short video, and hour-long video using Vicuna-7B based models.
}
\label{fig:demo}
\end{figure*}

\subsection{Main Results}
\noindent
\textbf{Results on Video-based Benchmarks.}
In Table~\ref{tab:main_video}, we provide a comparative evaluation of LLaMA-VID against various state-of-the-art methods across three zero-shot video QA benchmarks: MSVD-QA~\cite{msvd}, MSRVTT-QA~\cite{msrvtt}, and ActivityNet-QA~\cite{activitynet}.
Notably, the results are reported with only two tokens for each frame.
It is evident that LLaMA-VID, employing Vicuna-7B and Vicuna-13B as the LLMs, consistently delivers superior performance across all datasets. 
On the MSVD-QA and MSRVTT-QA datasets, it achieves the accuracy of 69.7\% and 57.7\% with Vicuna-7B, surpassing the previous leading approach~\cite{btadapter} with absolute gains of 2.2\% and 0.7\%, respectively.
As for the ActivityNet-QA dataset, LLaMA-VID attains top performance in accuracy and the highest score of 3.3.
In Table~\ref{tab:main_video_gen}, we also carry out experiments on the newly proposed video-based generative performance benchmark~\cite{videochatgpt}.
Our LLaMA-VID is validated to achieve the best performance across all the evaluation metrics, surpassing previous approaches by a large margin.
And we find that the performance can be further improved as the LLMs scale up.
In general, LLaMA-VID is demonstrated to bring robust performance on all benchmarks, validating its effectiveness and efficiency for video processing.

\vspace{1.0em}
\noindent
\textbf{Results on Image-based Benchmarks.}
As illustrated in Section~\ref{sec:sub_token}, LLaMA-VID not only efficiently represents long video, but also expands the upper limit of VLMs with an additional context token.
In Table~\ref{tab:main_img}, we perform comparisons with leading VLMs on 8 widely-adopted benchmarks.
It should be noted that we maintain the same training data and image resolution as in LLaVA-1.5~\cite{llava1.5} for fair comparisons.
It is evident that LLaMA-VID outperforms other leading methods across most of benchmarks with different LLMs.
In particular, with Vicuna-7B as the LLM, LLaMA-VID attains the best results across all the datasets and surpasses LLaVA-1.5 with significant gains in GQA, MME, and VizWiz, where the improvement reaches up to 2.3\%, 10.7, and 4.2\%, respectively.
With a larger Vicuna-13B as the LLM, LLaMA-VID also outperforms other methods in 6 benchmarks and achieves top-2 in the other datasets.
This demonstrates the generality of the proposed LLaMA-VID, which can be scaled up with a stronger foundation model.
In summary, LLaMA-VID is proven to push the upper bound of VLMs, especially in efficient settings.

\noindent
\textbf{Qualitative Results.}
In Figure~\ref{fig:demo}, we apply LLaMA-VID to different types of data, including single images, short videos, and long movies.
We represent each image with 577 tokens for single images and 2 tokens for videos.
LLaMA-VID demonstrates various capabilities with different inputs.
Specifically, for single images, it focuses on details and accurately recognizes the character without any text clues.
Moreover, it can also connect the image content to the plot of the game in multi-turn conversations.
Given a short trailer video, LLaMA-VID summarizes the overall plot and infers the movie name, producer, and the type of movie. 
As for a 3-hour movie, the proposed model adeptly describes the storyline and demonstrates plot-related reasoning and detailed understanding.

\subsection{Component-wise Analysis}~\label{sec:exp_abla}
In this subsection, we conduct ablation studies with input resolution 224 and 2 tokens for each image by default.
Here, we mainly perform experiments on image-based settings to investigate the effectiveness of each component.

\vspace{1.0em}
\noindent
\textbf{Generated Token Types.}
As illustrated in Figure~\ref{fig:main}, each image is represented with a context token and a content token in LLMs.
To validate the effectiveness of each part, we conduct experiments with different types of tokens in Table~\ref{tab:abla_token_type}. 
Without the context token, the compressed content token, which encodes each image with 1 token, cannot adjust to input instructions, leading to subpar performance.
Compared with a single content token, the instruction-guided context token results in significant gains across all datasets with only 1 token.
With both tokens for each image, the model achieves the best performance across all benchmarks.
It shows that both instruction cues in the context token and the image content itself in the content token are important.

\begin{table}[t]
 \centering
 \resizebox{0.49\textwidth}{12mm}{
\begin{tabular}{cc|cccc}
  \toprule
  {\em context} & {\em content} & {\bf GQA} & {\bf POPE} & {\bf SQA$^\text{I}$} & {\bf VQA$^\text{T}$}  \\
  \midrule
  \xmark & \cmark & 53.3 & 80.9 & 66.1 & 46.5 \\
  \cmark & \xmark & 54.3 & 82.4 & 67.7 & 48.3 \\
  \rowcolor{mygray}
  \cmark & \cmark & {\bf 55.5} & {\bf 83.1} & {\bf 68.8} & {\bf 49.0} \\
  \bottomrule
\end{tabular}
}
 \caption{Comparison with different token types.
We report results with 1 {\em context} token (if exists) and 1 {\em content} token.
 }
 \label{tab:abla_token_type}
\end{table}

\vspace{1.0em}
\noindent
\textbf{Generated Token Numbers.}
In Table~\ref{tab:abla_token_num}, we conduct experiments with different numbers of tokens for further investigation.
With an image size 224$\times$224, we set up experiments with $n$ content tokens, where $n=(224/14)^2=256$ for uncompressed settings in the first two rows.
The results clearly show that the context token consistently improves performance across different benchmarks with only 1 extra token.
When we compress the content token to $1/4$ with $n=64$, the performance drops about 1\% to 2\% but increases 1\% in SQA$^\text{I}$. 
Considering the extra efficient setting for hour-long videos, we compress the content token to $1/256$ with $n=1$ by default.
Compared to the original setting without context token, we can reduce the computational cost to $1/128$ with about 2\%-6\% performance drop, which is generally acceptable.
The linear increase in performance presents significant potential for token compression.
For instance, we can dynamically compress the content token to different numbers according to resource budget and content importance.
Interestingly, the model achieves peak performance in SQA$^\text{I}$ with only 2 tokens.
This could be attributed to the fact that problems in ScienceQA~\cite{scienceqa} focus more on visual-based reasoning rather than image details.
As demonstrated in Tables~\ref{tab:main_video} and~\ref{tab:main_video_gen}, with only 2 tokens for each image, LLaMA-VID still outperforms all previous work in different video-based benchmarks.
This makes it feasible to enable LLMs for hour-long video processing.

\begin{table}[t]
 \centering
\begin{tabular}{cc|cccc}
  \toprule
  {\em context} & {\em content} & {\bf GQA}  & {\bf POPE} & {\bf SQA$^\text{I}$} & {\bf VQA$^\text{T}$} \\
  \midrule
  0 & 256 & 61.9 & 85.5 & 67.5 & 53.0 \\
  \midrule
  1 & 256 & {\bf 63.0} & {\bf 86.6} & 67.7 & {\bf 53.8} \\
  1 & 64 & 60.8 & 85.1 & 68.7 & 52.3 \\
  1 & 16 & 58.2  & 83.1 & 67.4 & 50.8 \\
  1 & 4 & 56.2  & 83.5 & 68.7 & 49.1 \\
  \rowcolor{mygray}
  1 & 1 & 55.5  & 83.1 & {\bf 68.8} & 49.0 \\

  \bottomrule
\end{tabular}
 \caption{Comparison with different token numbers.
We report results with various numbers of {\em context} token and {\em content} token.
 }
 \label{tab:abla_token_num}
\end{table}

\begin{table}[t]
 \centering
\begin{tabular}{c|cccc}
  \toprule
  {\em text} & {\bf GQA}  & {\bf POPE} & {\bf SQA$^\text{I}$} & {\bf VQA$^\text{T}$} \\
  \midrule
  -- & 53.3  & 80.9 & 66.1 & 46.5 \\
  BERT & 54.1 & 80.8 & 67.9 & 48.1 \\
  \rowcolor{mygray}
  QFormer & {\bf 55.5} & {\bf 83.1} & {\bf 68.8} & {\bf 49.0} \\
  \bottomrule
\end{tabular}
  \caption{Comparison with different text decoders.
We report results with 1 {\em context} token (if exists) and 1 {\em content} token.
 }
 \label{tab:abla_text_vis}
\end{table}

\vspace{1.0em}
\noindent
\textbf{Text Decoder.}
As depicted in Figure~\ref{fig:main}, the text decoder plays an essential role in producing instruction-guided context cues.
Here, we further perform comparisons with different text decoders in Table~\ref{tab:abla_text_vis}.
We mainly instantiate the text decoder with two types of modules, namely BERT~\cite{devlin2018bert} and QFormer~\cite{instructblip}.
For BERT, we randomly initialize it as a cross-modality decoder and only retain the first two layers.
As for QFormer, we utilize the pre-trained modules and fix them for modality alignment.
Even with a simple 2-layer BERT, as shown in Table~\ref{tab:abla_text_vis}, the generated context token achieves significant gains in most of benchmarks.
This proves the effectiveness of the paradigm for context token generation.
With a pre-trained text decoder like QFormer, the model can be further enhanced and attains peak performance in all datasets with 2.2\% to 2.7\% significant gain.

\begin{figure}[t!]
\centering
\includegraphics[width=\linewidth]{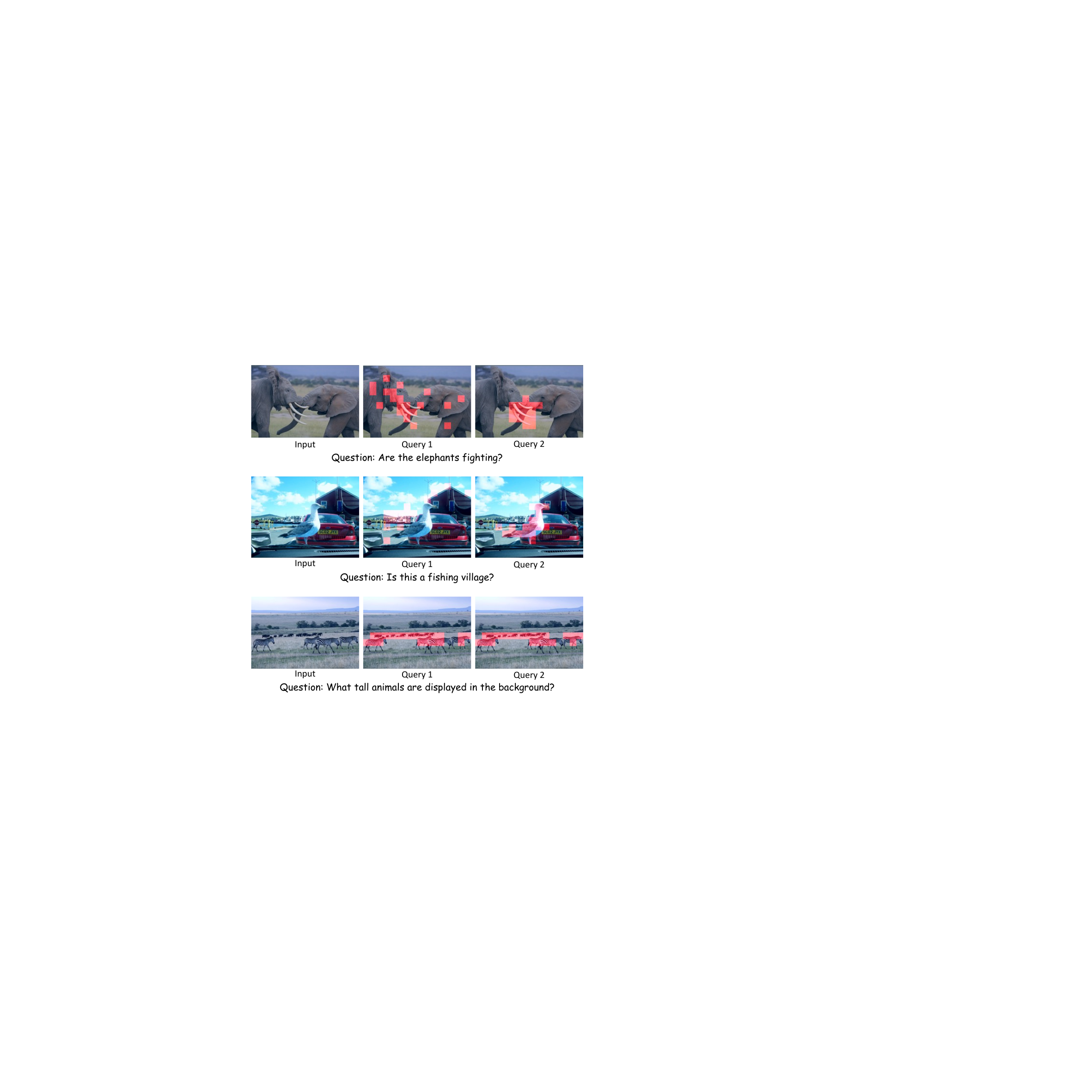} 
\caption{
High response areas with top scores to input question in Equation~\ref{equ:context_att}.
We present the response of first two queries in ${\mathbf Q}_t$.
Images are randomly sampled from VQA V2~\cite{vqav2} {\em test-dev} set.
}
\label{fig:heatmap}
\end{figure}

\vspace{1.0em}
\noindent
\textbf{Response in Context Attention.}
To more vividly explore the context attention, we visualize the high response areas with the top 20 scores in Figure~\ref{fig:heatmap}.
Specifically, we draw the normalized heatmap for the first two queries in ${\mathbf Q}_t$ before applying the $\text{Softmax}$ function, which is used to formulate context token in Equation~\ref{equ:context_att}.
As shown in Figure~\ref{fig:heatmap}, the text-guided query ${\mathbf Q}_t$ effectively focuses on important areas relevant to the input questions.
For example, in the second row, when inquiring whether the image depicts a fishing village, the query ${\mathbf Q}_t$ focuses more on buildings along the river and a seagull.
These are all typical characteristics to distinguish a fishing village in common sense.
Other examples also confirm that the designed context attention successfully achieves its goal of formulating the context token under instruction guidance.
More visualizations can be found in the attached {\em supplementary material}.

\section{Conclusion}
We have introduced LLaMA-VID, a simple yet effective token generation approach for VLMs.
The central concept behind LLaMA-VID is to represent an image with the context token and the content token.
In particular, the context token is generated according to input instructions, and the content token is produced based on the image content.
Depending on the budget, the content token can be compressed to one token or expressed without compression.
It allows us to represent a single image with preserved details and efficiently encode each video frame with only two tokens.
Moreover, we have constructed an instruction dataset for hour-long video understanding.
Our experiments on several video- and image-based benchmarks prove the superiority of our method.
We hope that LLaMA-VID can serve as a strong benchmark for efficient visual representation.

\appendix

\section{Experimental Details}
In this section, we delve into the experimental details of the proposed LLaMA-VID framework.
Generally, we adopt a similar training strategy with that in~\cite{llava1.5}, except that we freeze the proposed text decoder during pretraining and subsequently optimize it during the finetuning phase.
As outlined in Table~\ref{tab:train_detail}, we employ customized training settings for the distinct stages illustrated in Figure~\ref{fig:data} of the main paper. 
Specifically, during the modality alignment phase (stage 1), we keep the text decoder fixed and limit the maximum token length to 2K. 
Moving on to instruction tuning (stage 2), we unfreeze and optimize the text decoder, accommodating a larger maximum token length of 2K to facilitate video tuning.
If stage 3 is adopted for long video tuning, we once again freeze the text decoder to conserve memory resources while increasing the maximum token length to a substantial 64K, catering to the demands of hour-long movie content. 
Detailed training settings are further explicated in Table~\ref{tab:train_detail}.

\begin{table}[h]
\centering
\begin{tabular}{l| c c c}
\toprule
Settings & Stage 1 & Stage 2 & Stage 3 \\
\midrule
Batch size & 256 & 128 & 8 \\
Learning rate & 1e-3 & 2e-5 & 2e-5 \\
Learning schedule & \multicolumn{3}{c}{Cosine decay} \\
Warmup ratio & \multicolumn{3}{c}{0.03} \\
Weight decay & \multicolumn{3}{c}{0} \\
Epoch & \multicolumn{3}{c}{1} \\
Optimizer & \multicolumn{3}{c}{AdamW} \\
DeepSpeed stage & \multicolumn{3}{c}{2} \\
Vision encoder & \multicolumn{3}{c}{Freeze} \\
Text decoder & Freeze & Open & Freeze \\
Max token & 2048 & 2048 & 65536 \\
\bottomrule
\end{tabular}
\caption{
Training settings of LLaMA-VID.
}
\label{tab:train_detail}
\end{table}

\section{Instruction Tuning Details}

\noindent
\textbf{Context Extension.} 
To support hour-long video understanding, we conduct context extension of the language model in LLaMA-VID to accommodate inputs of up to 64K tokens.
In line with prior research on long context LLMs~\cite{position_interpolation,longlora}, we employ position interpolation techniques~\cite{position_interpolation} to scale the rotary position encoding~\cite{rope} from 4K to 64K, thereby enabling the processing of extended sequences. 
Subsequent to this adaptation, we perform supervised fine-tuning using Long-VideoQA, a dataset specifically designed for long-duration video instruction-following tasks.

\vspace{1.0em}
\noindent
\textbf{Long-VideoQA Dataset.} 
To improve the ability for understanding hour-long videos, we develop a specialized dataset for supervised fine-tuning purposes, named Long-VideoQA. 
The Long-VideoQA dataset comprises 15K question-and-answer (QA) pairs, with 9K of these pairs derived from movie scenes and the remaining 6K pairs sourced from LongAlpaca~\cite{longlora}. 
This dataset is designed to improve the performance of models on long-duration videos by providing them with relevant and diverse instruction-following scenarios.
Inspired by previous works on image and short video instruction-following data collection~\cite{llava,videochat,videochatgpt}, we leverage advanced language models like GPT-4~\cite{GPT4} and Claude-2~\cite{Claude2} as the strong teacher, to create instruction-following data that incorporates video content. 
To effectively prompt these language models, we utilized two types of symbolic representations to encode the video content:
\begin{itemize}
    \item \textit{Synopses}. It offers comprehensive narratives detailing the plot of a movie. These synopses give a broad overview of the storyline, encompassing major events and character developments.

    \item \textit{Scripts}. It provides a more granular representation, including the storyline, character dialogues, and specific actions. Scripts convey the content in a format that closely follows the actual sequence of scenes and dialogues.
\end{itemize}

\begin{figure*}[t]
\centering
\includegraphics[width=\linewidth]{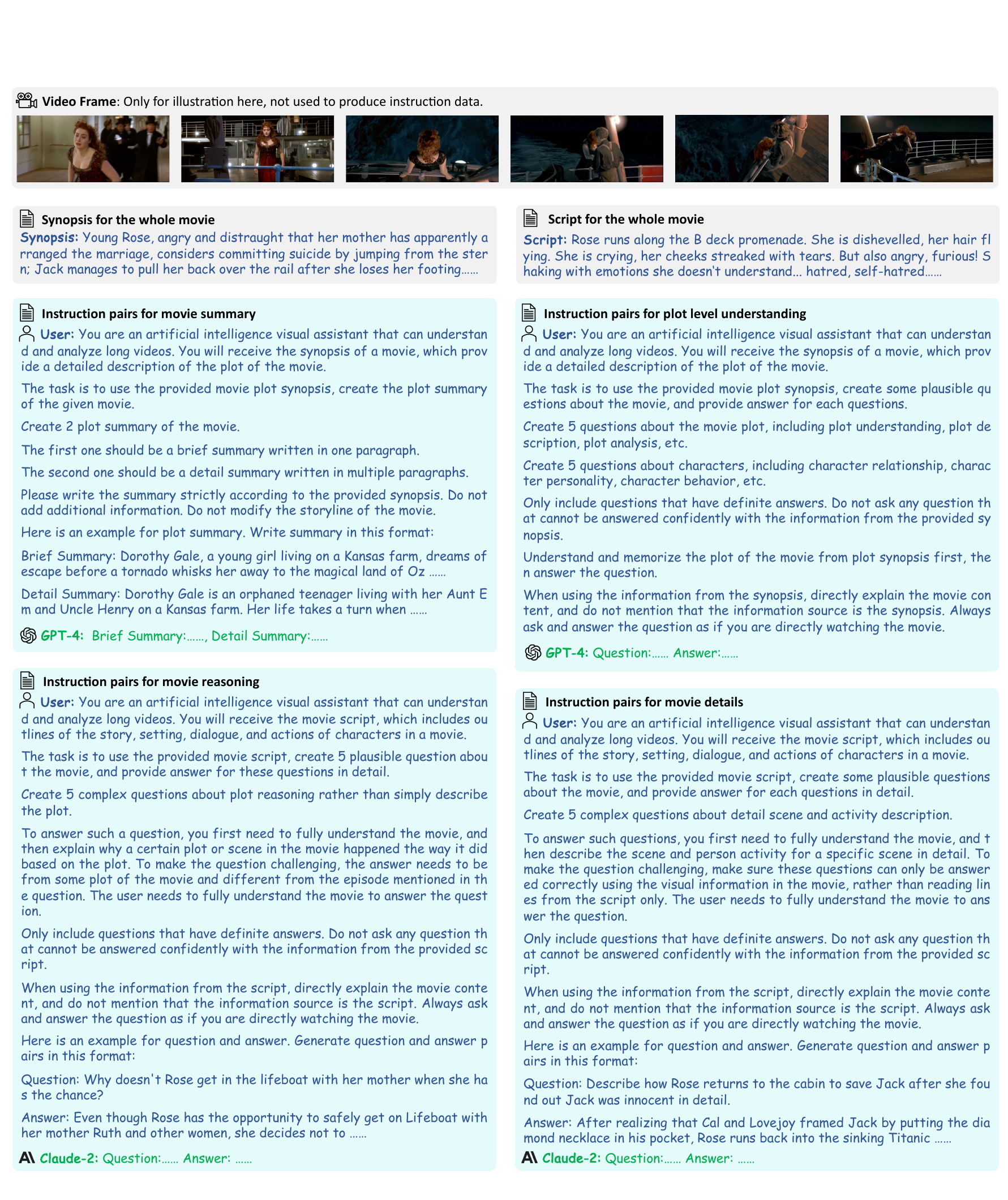} 
\caption{
Details in constructing instruction pairs for the movie Titanic.
It is detailed version of that in Figure~\ref{fig:long_movie} of the main paper.
}
\label{fig:long_movie_appendix}
\end{figure*}

By utilizing movies from MovieNet~\cite{movienet} along with their corresponding synopses and scripts, we were able to create a rich dataset. 
It serves as a foundation for finetuning models to understand and respond to instructions in the context of long-form video content. 
As illustrated in Figure~\ref{fig:long_movie_appendix}, we prompt GPT-4/Claude2 to generate three types of instruction-following data:
\begin{itemize}
    \item \textit{Video summary.} 
    Utilizing synopses as prompts, we engage GPT-4 to generate two types of summaries for each movie: one brief summary and one detailed summary. This approach is designed to equip the model with a comprehensive understanding of the video content, providing both a concise encapsulation and an in-depth exposition of the narrative.
    The brief summary offers a quick snapshot of the movie's overarching plot, while the detailed summary delves into the nuances, including character arcs, thematic elements, and key plot points. 
    \item \textit{Plot-level understanding.} 
    Leveraging the synopses as input prompts, we employ GPT-4 to generate both plot-related and character-related QA pairs. 
    This process is tailored to yield a deeper insight into the movie's narrative structure and the intricacies of character development.
    The plot-related QA pairs focus on the events, settings, and conflicts within the movie, enabling the model to grasp the story progression and the causal relationships between different events.
    Simultaneously, the character-related QA pairs delve into the personalities, motivations, and relationships of the characters, offering a nuanced perspective on how each character influences and is influenced by the unfolding story.
    In this way, we ensure that the model acquires a well-rounded comprehension of both the plot dynamics and the character profiles, which is crucial for accurately understanding and responding to complex instructions related to hour-long video content.
    \item \textit{Detail description and reasoning.} 
    By inputting the extensive movie scripts into Claude-2, we generate two distinct sets of data for each film: 5 plot-related reasoning pairs and 5 detail-related descriptions. 
    This methodology serves to enhance the model's capacity for learning and understanding fine-grained details.
    The plot-related reasoning pairs are crafted to challenge the model ability to make inferences and understand the logical flow of the narrative. 
    These pairs require the model to connect different plot points, reason about cause and effect, and predict outcomes based on earlier events in the story.
    On the other hand, the detail-related descriptions focus on capturing the subtleties within the movie, such as specific descriptions of scenes, nuanced character expressions, or particular dialogue exchanges. 
    These descriptions are crucial for the model to learn about intricacies that are often pivotal for a comprehensive understanding of video content.
        
\end{itemize}

\section{Additional Analyses}
We provide more examples for long videos and images, as well as the visualization for response in context attention.

\begin{figure*}[t]
\centering
\includegraphics[width=\linewidth]{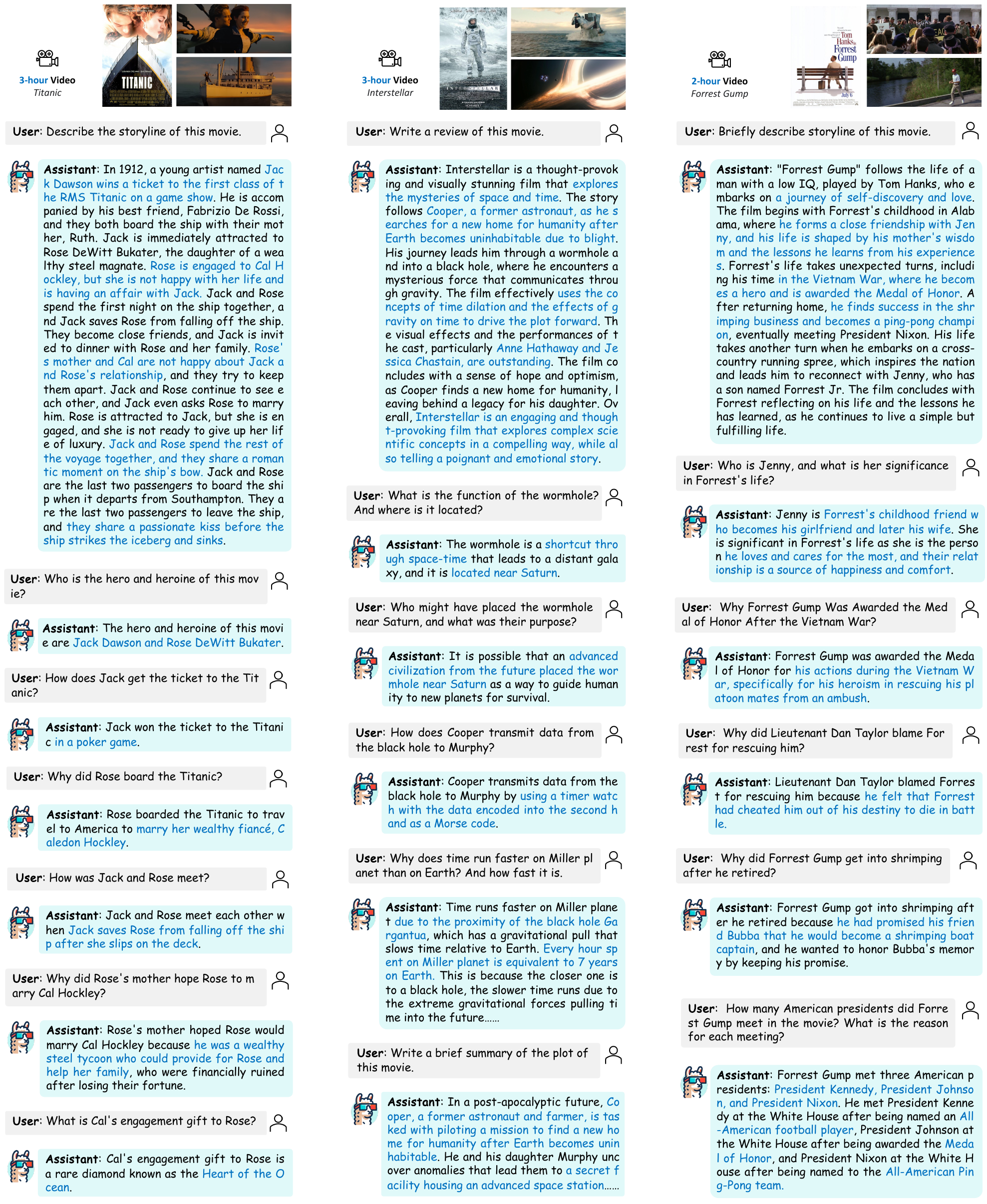} 
\caption{
More examples of LLaMA-VID with hour-long movies using Vicuna-7B based models.
}
\label{fig:demo_movie_appendix}
\end{figure*}

\begin{figure*}[t]
\centering
\includegraphics[width=\linewidth]{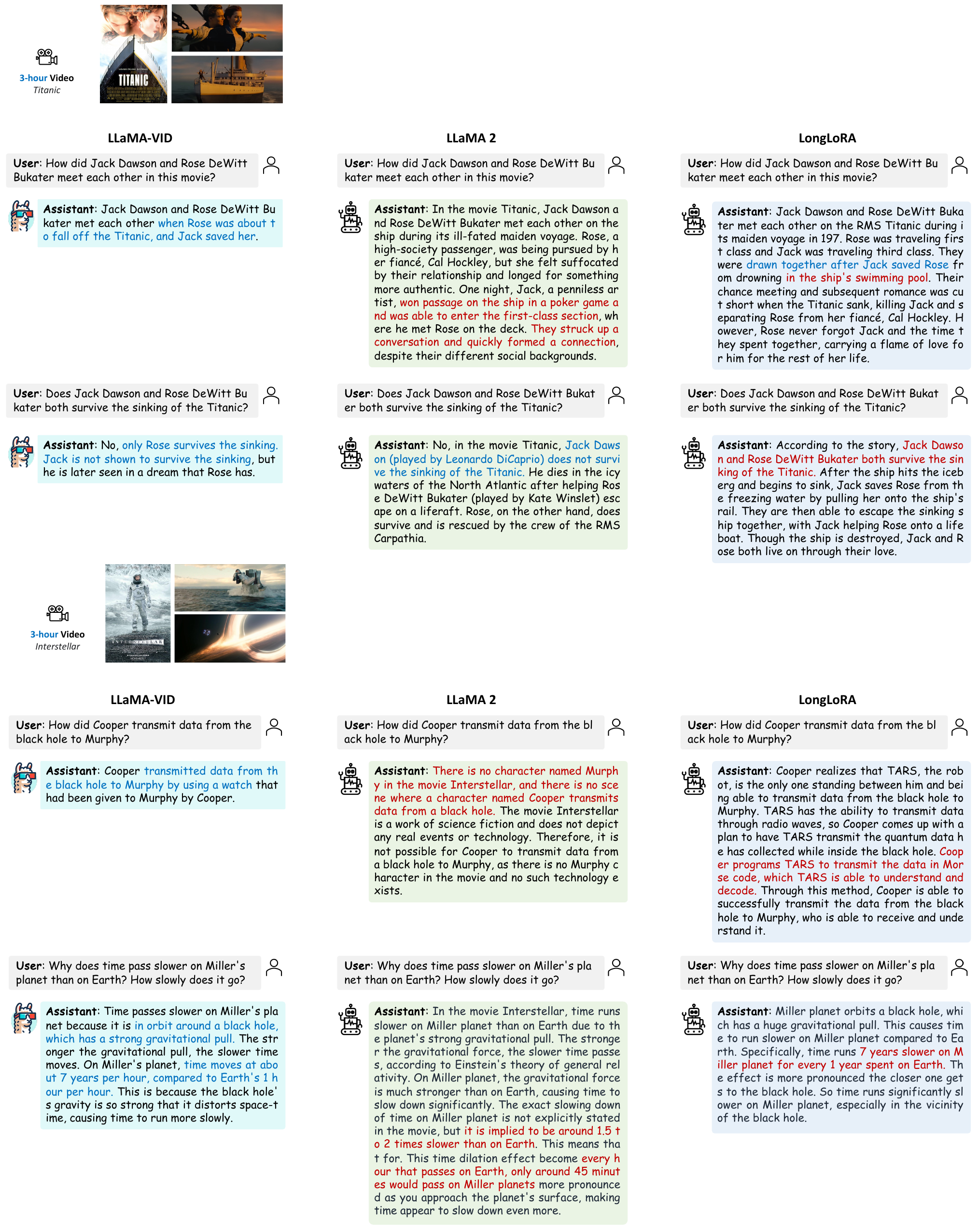} 
\caption{
Comparisons with LLaMA 2~\cite{llama2} and LongLoRA~\cite{longlora} given the same questions.
}
\label{fig:movie_compare}
\end{figure*}

\vspace{1.0em}
\noindent
\textbf{Examples of Long Videos.} 
In Figure~\ref{fig:demo_movie_appendix}, we further demonstrate the effectiveness of LLaMA-VID by interacting with a broader range of hour-long movies, including genres like Romance, Adventure, and Sci-Fi. 
This diverse selection of films allows us to evaluate the performance across different narrative structures and thematic elements.
It demonstrates a proficient ability to summarize storylines and engage in plot-related reasoning by synthesizing information from both video frames and subtitles. 
In particular, we compare with LLaMA 2~\cite{llama2} and LongLoRA~\cite{longlora} by feeding the movie name and all subtitles, respectively.
As depicted in Figure~\ref{fig:movie_compare}, the result reveals that LLaMA-VID outperforms its counterparts in tasks involving character understanding and plot-related reasoning.
The nuanced understanding of complex narratives and the integration of multi-modal data underscore its potential for advanced applications in video analysis and interaction.

\begin{figure*}[t]
\centering
\includegraphics[width=\linewidth]{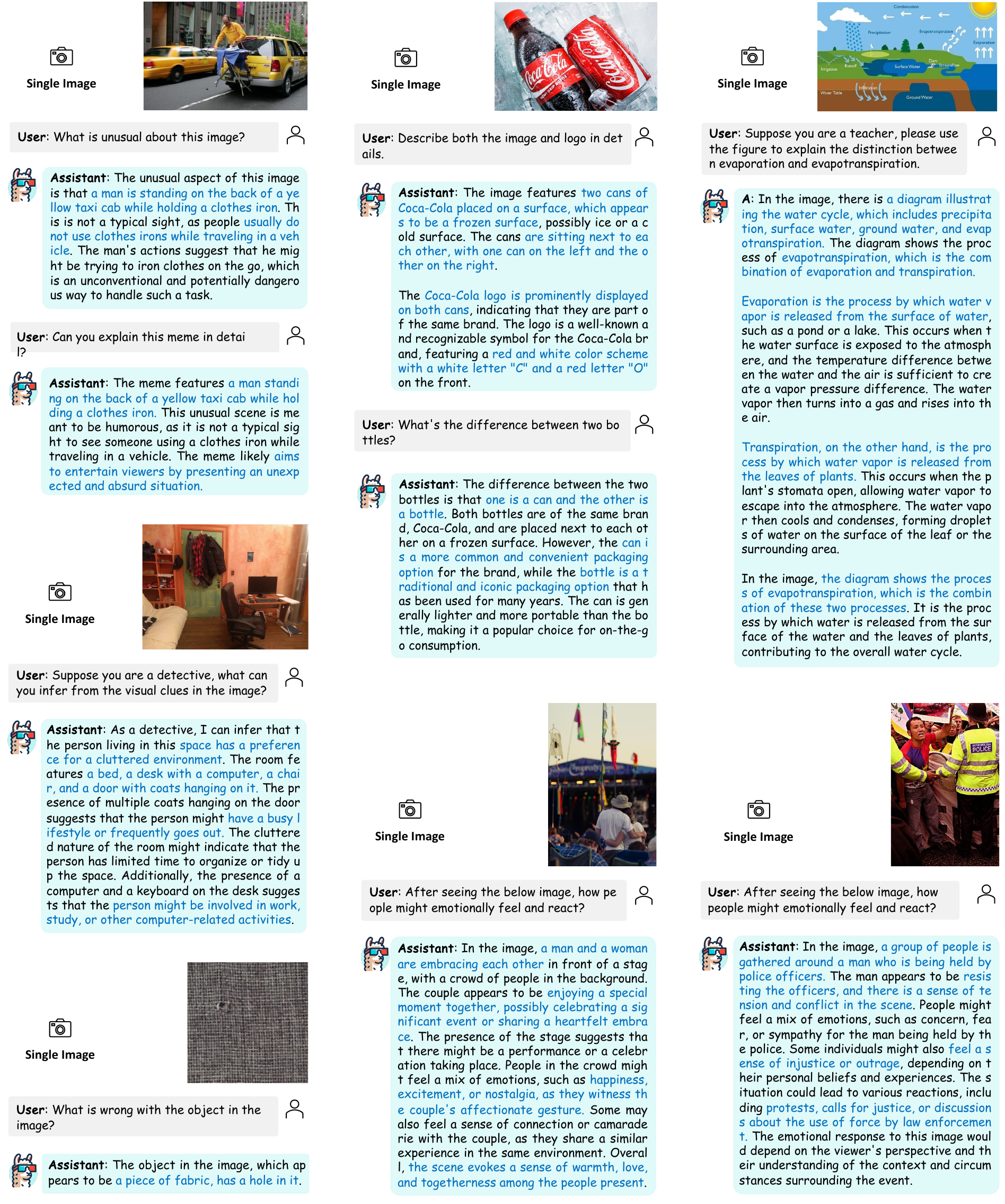} 
\caption{
More examples of LLaMA-VID with single image using Vicuna-7B based models.
Images are sampled from~\cite{yang2023dawn}.
}
\label{fig:demo_fig_appendix}
\end{figure*}

\vspace{1.0em}
\noindent
\textbf{Examples of Images.} 
In Figure~\ref{fig:demo_fig_appendix}, we present additional interactions with LLaMA-VID, showcasing the model's adeptness in knowledge-based perception and reasoning. 
The examples illustrate the model's capability to discern and interpret object details within a given environment and to utilize environmental cues to effectively respond to user inquiries.
For instance, when provided with a description or an image of the surrounding environment, LLaMA-VID is able to identify specific objects and their attributes. 
It can then leverage this information to answer questions posed by the user that may relate to object functions, spatial relationships, or contextual relevance.
This performance is indicative of the sophisticated understanding.
It is not limited to mere recognition but extends to a deeper cognitive level where it can process and integrate environmental information to engage in informed and accurate dialogue. 
Such demonstrations reinforce the potential of LLaMA-VID as a tool for complex interaction involving visual data and natural language processing.

\begin{figure*}[t]
\centering
\includegraphics[width=0.95\linewidth]{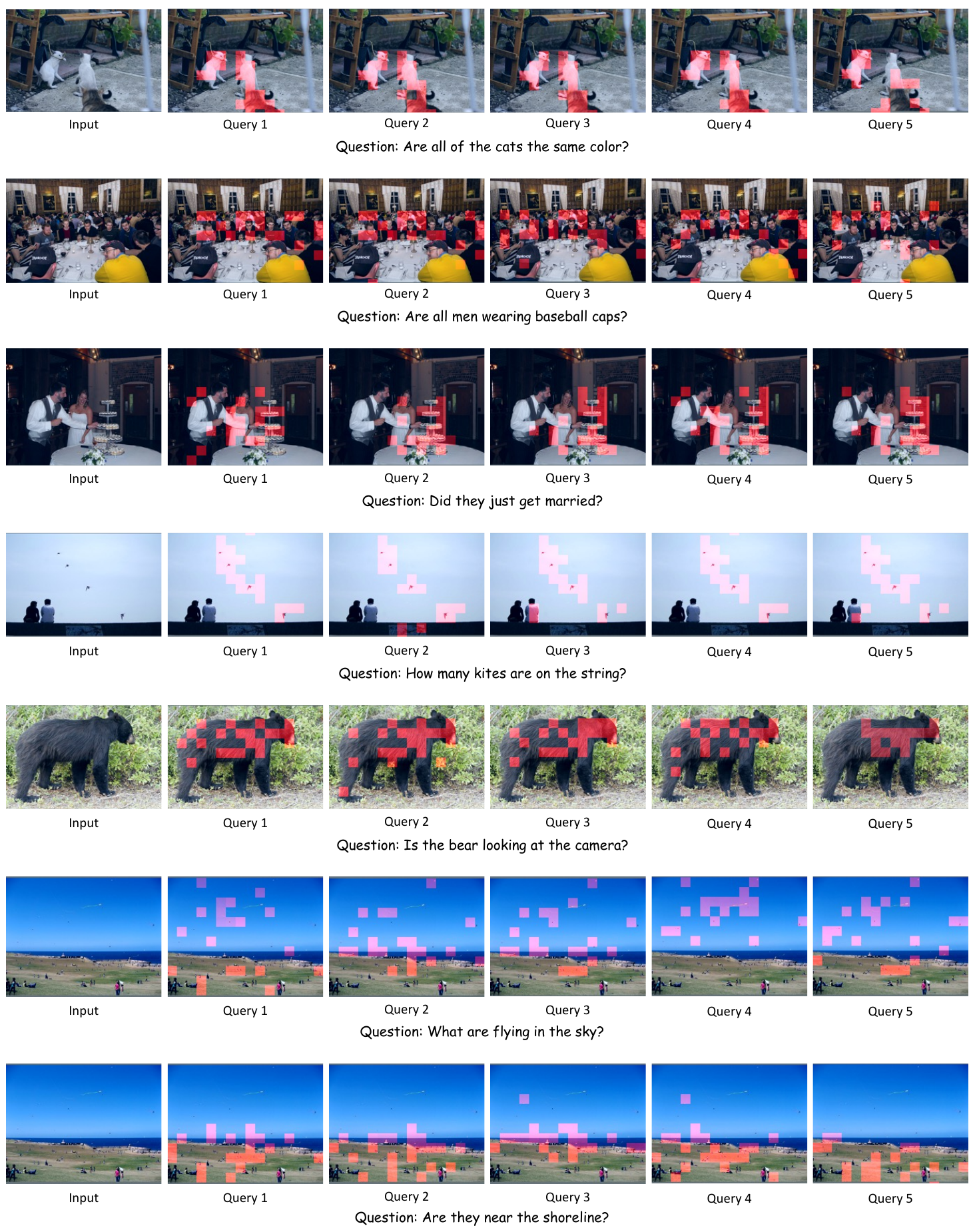} 
\caption{
High response areas with top scores to input question in Equation~\ref{equ:context_att} of the main paper.
We present the response of first five queries in ${\mathbf Q}_t$.
Images are randomly sampled from VQA V2~\cite{vqav2} {\em test-dev} set.
}
\label{fig:heatmap_appendix}
\end{figure*}

\vspace{1.0em}
\noindent
\textbf{Response in Context Attention.} 
In Figure~\ref{fig:heatmap_appendix}, we extend the visualization of the attention mechanism by presenting additional results of the response in context attention. 
This visualization is in alignment with that in Figure~\ref{fig:heatmap} of the main paper, where we highlight the areas with the highest attention scores.
Specifically, we focus on the top 20 scores while retaining the first five queries in ${\mathbf Q}_t$.
These visualizations serve to underline the effectiveness of the text-guided query ${\mathbf Q}_t$ on areas that are most pertinent to input questions. 
It is dynamically adjusted based on the content of input questions, demonstrating its ability to discern and prioritize different regions depending on the context.
For example, in the last two rows of Figure~\ref{fig:heatmap_appendix}, where different questions are posed regarding the same image, LLaMA-VID adjusts its focus accordingly, directing its attention to specific regions that are relevant to each question. 
This capability signifies a sophisticated level of contextual understanding and adaptability, showcasing the potential of LLaMA-VID in processing complex visual and textual inputs.

\clearpage

{
    \small
    \bibliographystyle{ieeenat_fullname}
    \bibliography{arxiv}

\begin{thebibliography}{65}
\providecommand{\natexlab}[1]{#1}
\providecommand{\url}[1]{\texttt{#1}}
\expandafter\ifx\csname urlstyle\endcsname\relax
  \providecommand{\doi}[1]{doi: #1}\else
  \providecommand{\doi}{doi: \begingroup \urlstyle{rm}\Url}\fi

\bibitem[Sha(2023)]{ShareGPT}
Sharegpt.
\newblock \url{https://sharegpt.com/}, 2023.

\bibitem[Alayrac et~al.(2022)Alayrac, Donahue, Luc, Miech, Barr, Hasson, Lenc,
  Mensch, Millican, Reynolds, et~al.]{flamingo}
Jean-Baptiste Alayrac, Jeff Donahue, Pauline Luc, Antoine Miech, Iain Barr,
  Yana Hasson, Karel Lenc, Arthur Mensch, Katherine Millican, Malcolm Reynolds,
  et~al.
\newblock Flamingo: a visual language model for few-shot learning.
\newblock In \emph{NeurIPS}, 2022.

\bibitem[Anthropic(2023)]{Claude2}
Anthropic.
\newblock Claude 2.
\newblock \url{https://www.anthropic.com/index/claude-2}, 2023.

\bibitem[Bai et~al.(2023)Bai, Bai, Yang, Wang, Tan, Wang, Lin, Zhou, and
  Zhou]{bai2023qwen}
Jinze Bai, Shuai Bai, Shusheng Yang, Shijie Wang, Sinan Tan, Peng Wang, Junyang
  Lin, Chang Zhou, and Jingren Zhou.
\newblock Qwen-vl: A frontier large vision-language model with versatile
  abilities.
\newblock \emph{arXiv:2308.12966}, 2023.

\bibitem[Bain et~al.(2021)Bain, Nagrani, Varol, and Zisserman]{webvid}
Max Bain, Arsha Nagrani, G{\"u}l Varol, and Andrew Zisserman.
\newblock Frozen in time: A joint video and image encoder for end-to-end
  retrieval.
\newblock In \emph{ICCV}, 2021.

\bibitem[Brown et~al.(2020)Brown, Mann, Ryder, Subbiah, Kaplan, Dhariwal,
  Neelakantan, Shyam, Sastry, Askell, et~al.]{brown2020language}
Tom Brown, Benjamin Mann, Nick Ryder, Melanie Subbiah, Jared~D Kaplan, Prafulla
  Dhariwal, Arvind Neelakantan, Pranav Shyam, Girish Sastry, Amanda Askell,
  et~al.
\newblock Language models are few-shot learners.
\newblock In \emph{NeurIPS}, 2020.

\bibitem[Caba~Heilbron et~al.(2015)Caba~Heilbron, Escorcia, Ghanem, and
  Carlos~Niebles]{activitynet}
Fabian Caba~Heilbron, Victor Escorcia, Bernard Ghanem, and Juan Carlos~Niebles.
\newblock Activitynet: A large-scale video benchmark for human activity
  understanding.
\newblock In \emph{CVPR}, 2015.

\bibitem[Chen and Dolan(2011)]{msvd}
David Chen and William~B Dolan.
\newblock Collecting highly parallel data for paraphrase evaluation.
\newblock In \emph{ACL}, 2011.

\bibitem[Chen et~al.(2023{\natexlab{a}})Chen, Zhang, Zeng, Zhang, Zhu, and
  Zhao]{chen2023shikra}
Keqin Chen, Zhao Zhang, Weili Zeng, Richong Zhang, Feng Zhu, and Rui Zhao.
\newblock Shikra: Unleashing multimodal llm's referential dialogue magic.
\newblock \emph{arXiv:2306.15195}, 2023{\natexlab{a}}.

\bibitem[Chen et~al.(2023{\natexlab{b}})Chen, Wong, Chen, and
  Tian]{position_interpolation}
Shouyuan Chen, Sherman Wong, Liangjian Chen, and Yuandong Tian.
\newblock Extending context window of large language models via positional
  interpolation.
\newblock \emph{arXiv:2306.15595}, 2023{\natexlab{b}}.

\bibitem[Chen et~al.(2015)Chen, Fang, Lin, Vedantam, Gupta, Doll{\'a}r, and
  Zitnick]{cococap}
Xinlei Chen, Hao Fang, Tsung-Yi Lin, Ramakrishna Vedantam, Saurabh Gupta, Piotr
  Doll{\'a}r, and C~Lawrence Zitnick.
\newblock Microsoft coco captions: Data collection and evaluation server.
\newblock \emph{arXiv:1504.00325}, 2015.

\bibitem[Chen et~al.(2023{\natexlab{c}})Chen, Qian, Tang, Lai, Liu, Han, and
  Jia]{longlora}
Yukang Chen, Shengju Qian, Haotian Tang, Xin Lai, Zhijian Liu, Song Han, and
  Jiaya Jia.
\newblock Longlora: Efficient fine-tuning of long-context large language
  models.
\newblock \emph{arXiv:2309.12307}, 2023{\natexlab{c}}.

\bibitem[Chiang et~al.(2023)Chiang, Li, Lin, Sheng, Wu, Zhang, Zheng, Zhuang,
  Zhuang, Gonzalez, Stoica, and Xing]{vicuna}
Wei-Lin Chiang, Zhuohan Li, Zi Lin, Ying Sheng, Zhanghao Wu, Hao Zhang, Lianmin
  Zheng, Siyuan Zhuang, Yonghao Zhuang, Joseph~E. Gonzalez, Ion Stoica, and
  Eric~P. Xing.
\newblock Vicuna: An open-source chatbot impressing gpt-4 with 90\%* chatgpt
  quality.
\newblock \url{https://lmsys.org/blog/2023-03-30-vicuna/}, 2023.

\bibitem[Dai et~al.(2023)Dai, Li, Li, Tiong, Zhao, Wang, Li, Fung, and
  Hoi]{instructblip}
Wenliang Dai, Junnan Li, Dongxu Li, Anthony Meng~Huat Tiong, Junqi Zhao,
  Weisheng Wang, Boyang Li, Pascale Fung, and Steven Hoi.
\newblock Instructblip: Towards general-purpose vision-language models with
  instruction tuning.
\newblock \emph{arXiv:2305.06500}, 2023.

\bibitem[Devlin et~al.(2018)Devlin, Chang, Lee, and Toutanova]{devlin2018bert}
Jacob Devlin, Ming-Wei Chang, Kenton Lee, and Kristina Toutanova.
\newblock Bert: Pre-training of deep bidirectional transformers for language
  understanding.
\newblock \emph{arXiv:1810.04805}, 2018.

\bibitem[Dosovitskiy et~al.(2021)Dosovitskiy, Beyer, Kolesnikov, Weissenborn,
  Zhai, Unterthiner, Dehghani, Minderer, Heigold, Gelly, et~al.]{vit}
Alexey Dosovitskiy, Lucas Beyer, Alexander Kolesnikov, Dirk Weissenborn,
  Xiaohua Zhai, Thomas Unterthiner, Mostafa Dehghani, Matthias Minderer, Georg
  Heigold, Sylvain Gelly, et~al.
\newblock An image is worth 16x16 words: Transformers for image recognition at
  scale.
\newblock In \emph{ICLR}, 2021.

\bibitem[Fang et~al.(2023)Fang, Wang, Xie, Sun, Wu, Wang, Huang, Wang, and
  Cao]{evaclip}
Yuxin Fang, Wen Wang, Binhui Xie, Quan Sun, Ledell Wu, Xinggang Wang, Tiejun
  Huang, Xinlong Wang, and Yue Cao.
\newblock Eva: Exploring the limits of masked visual representation learning at
  scale.
\newblock In \emph{CVPR}, 2023.

\bibitem[Fu et~al.(2023)Fu, Chen, Shen, Qin, Zhang, Lin, Qiu, Lin, Yang, Zheng,
  et~al.]{mme}
Chaoyou Fu, Peixian Chen, Yunhang Shen, Yulei Qin, Mengdan Zhang, Xu Lin,
  Zhenyu Qiu, Wei Lin, Jinrui Yang, Xiawu Zheng, et~al.
\newblock Mme: A comprehensive evaluation benchmark for multimodal large
  language models.
\newblock \emph{arXiv:2306.13394}, 2023.

\bibitem[Goyal et~al.(2017)Goyal, Khot, Summers-Stay, Batra, and Parikh]{vqav2}
Yash Goyal, Tejas Khot, Douglas Summers-Stay, Dhruv Batra, and Devi Parikh.
\newblock Making the v in vqa matter: Elevating the role of image understanding
  in visual question answering.
\newblock In \emph{CVPR}, 2017.

\bibitem[Gurari et~al.(2018)Gurari, Li, Stangl, Guo, Lin, Grauman, Luo, and
  Bigham]{vizwiz}
Danna Gurari, Qing Li, Abigale~J Stangl, Anhong Guo, Chi Lin, Kristen Grauman,
  Jiebo Luo, and Jeffrey~P Bigham.
\newblock Vizwiz grand challenge: Answering visual questions from blind people.
\newblock In \emph{CVPR}, 2018.

\bibitem[Huang et~al.(2020)Huang, Xiong, Rao, Wang, and Lin]{movienet}
Qingqiu Huang, Yu Xiong, Anyi Rao, Jiaze Wang, and Dahua Lin.
\newblock Movienet: A holistic dataset for movie understanding.
\newblock In \emph{ECCV}, 2020.

\bibitem[Hudson and Manning(2019)]{gqa}
Drew~A Hudson and Christopher~D Manning.
\newblock Gqa: A new dataset for real-world visual reasoning and compositional
  question answering.
\newblock In \emph{CVPR}, 2019.

\bibitem[IDEFICS(2023)]{IDEFICS}
IDEFICS.
\newblock Introducing idefics: An open reproduction of state-of-the-art visual
  language model.
\newblock \url{https://huggingface.co/blog/idefics}, 2023.

\bibitem[Jia et~al.(2021)Jia, Yang, Xia, Chen, Parekh, Pham, Le, Sung, Li, and
  Duerig]{ALIGN}
Chao Jia, Yinfei Yang, Ye Xia, Yi-Ting Chen, Zarana Parekh, Hieu Pham, Quoc Le,
  Yun-Hsuan Sung, Zhen Li, and Tom Duerig.
\newblock Scaling up visual and vision-language representation learning with
  noisy text supervision.
\newblock In \emph{ICML}, 2021.

\bibitem[Kazemzadeh et~al.(2014)Kazemzadeh, Ordonez, Matten, and
  Berg]{referitgame}
Sahar Kazemzadeh, Vicente Ordonez, Mark Matten, and Tamara Berg.
\newblock Referitgame: Referring to objects in photographs of natural scenes.
\newblock In \emph{EMNLP}, 2014.

\bibitem[Krishna et~al.(2017)Krishna, Zhu, Groth, Johnson, Hata, Kravitz, Chen,
  Kalantidis, Li, Shamma, et~al.]{vg}
Ranjay Krishna, Yuke Zhu, Oliver Groth, Justin Johnson, Kenji Hata, Joshua
  Kravitz, Stephanie Chen, Yannis Kalantidis, Li-Jia Li, David~A Shamma, et~al.
\newblock Visual genome: Connecting language and vision using crowdsourced
  dense image annotations.
\newblock \emph{IJCV}, 2017.

\bibitem[Lai et~al.(2023)Lai, Tian, Chen, Li, Yuan, Liu, and Jia]{lai2023lisa}
Xin Lai, Zhuotao Tian, Yukang Chen, Yanwei Li, Yuhui Yuan, Shu Liu, and Jiaya
  Jia.
\newblock Lisa: Reasoning segmentation via large language model.
\newblock \emph{arXiv:2308.00692}, 2023.

\bibitem[Li et~al.(2023{\natexlab{a}})Li, Wang, Wang, Ge, Ge, and Shan]{seed}
Bohao Li, Rui Wang, Guangzhi Wang, Yuying Ge, Yixiao Ge, and Ying Shan.
\newblock Seed-bench: Benchmarking multimodal llms with generative
  comprehension.
\newblock \emph{arXiv:2307.16125}, 2023{\natexlab{a}}.

\bibitem[Li et~al.(2023{\natexlab{b}})Li, Li, Savarese, and Hoi]{blip2}
Junnan Li, Dongxu Li, Silvio Savarese, and Steven Hoi.
\newblock Blip-2: Bootstrapping language-image pre-training with frozen image
  encoders and large language models.
\newblock \emph{arXiv:2301.12597}, 2023{\natexlab{b}}.

\bibitem[Li et~al.(2023{\natexlab{c}})Li, He, Wang, Li, Wang, Luo, Wang, Wang,
  and Qiao]{videochat}
KunChang Li, Yinan He, Yi Wang, Yizhuo Li, Wenhai Wang, Ping Luo, Yali Wang,
  Limin Wang, and Yu Qiao.
\newblock Videochat: Chat-centric video understanding.
\newblock \emph{arXiv:2305.06355}, 2023{\natexlab{c}}.

\bibitem[Li et~al.(2023{\natexlab{d}})Li, Du, Zhou, Wang, Zhao, and Wen]{pope}
Yifan Li, Yifan Du, Kun Zhou, Jinpeng Wang, Wayne~Xin Zhao, and Ji-Rong Wen.
\newblock Evaluating object hallucination in large vision-language models.
\newblock \emph{arXiv:2305.10355}, 2023{\natexlab{d}}.

\bibitem[Liu et~al.(2023{\natexlab{a}})Liu, Li, Li, and Lee]{llava1.5}
Haotian Liu, Chunyuan Li, Yuheng Li, and Yong~Jae Lee.
\newblock Improved baselines with visual instruction tuning.
\newblock \emph{arXiv:2310.03744}, 2023{\natexlab{a}}.

\bibitem[Liu et~al.(2023{\natexlab{b}})Liu, Li, Wu, and Lee]{llava}
Haotian Liu, Chunyuan Li, Qingyang Wu, and Yong~Jae Lee.
\newblock Visual instruction tuning.
\newblock In \emph{NeruIPS}, 2023{\natexlab{b}}.

\bibitem[Liu et~al.(2023{\natexlab{c}})Liu, Li, Ge, Shan, Li, and
  Li]{btadapter}
Ruyang Liu, Chen Li, Yixiao Ge, Ying Shan, Thomas~H Li, and Ge Li.
\newblock One for all: Video conversation is feasible without video instruction
  tuning.
\newblock \emph{arXiv:2309.15785}, 2023{\natexlab{c}}.

\bibitem[Liu et~al.(2019)Liu, Ott, Goyal, Du, Joshi, Chen, Levy, Lewis,
  Zettlemoyer, and Stoyanov]{liu2019roberta}
Yinhan Liu, Myle Ott, Naman Goyal, Jingfei Du, Mandar Joshi, Danqi Chen, Omer
  Levy, Mike Lewis, Luke Zettlemoyer, and Veselin Stoyanov.
\newblock Roberta: A robustly optimized bert pretraining approach.
\newblock \emph{arXiv:1907.11692}, 2019.

\bibitem[Liu et~al.(2023{\natexlab{d}})Liu, Duan, Zhang, Li, Zhang, Zhao, Yuan,
  Wang, He, Liu, et~al.]{mmbench}
Yuan Liu, Haodong Duan, Yuanhan Zhang, Bo Li, Songyang Zhang, Wangbo Zhao, Yike
  Yuan, Jiaqi Wang, Conghui He, Ziwei Liu, et~al.
\newblock Mmbench: Is your multi-modal model an all-around player?
\newblock \emph{arXiv:2307.06281}, 2023{\natexlab{d}}.

\bibitem[Lu et~al.(2022)Lu, Mishra, Xia, Qiu, Chang, Zhu, Tafjord, Clark, and
  Kalyan]{scienceqa}
Pan Lu, Swaroop Mishra, Tanglin Xia, Liang Qiu, Kai-Wei Chang, Song-Chun Zhu,
  Oyvind Tafjord, Peter Clark, and Ashwin Kalyan.
\newblock Learn to explain: Multimodal reasoning via thought chains for science
  question answering.
\newblock In \emph{NeurIPS}, 2022.

\bibitem[Luo et~al.(2023)Luo, Zhao, Yang, Dong, Qiu, Lu, Wang, and
  Wei]{luo2023valley}
Ruipu Luo, Ziwang Zhao, Min Yang, Junwei Dong, Minghui Qiu, Pengcheng Lu, Tao
  Wang, and Zhongyu Wei.
\newblock Valley: Video assistant with large language model enhanced ability.
\newblock \emph{arXiv:2306.07207}, 2023.

\bibitem[Maaz et~al.(2023)Maaz, Rasheed, Khan, and Khan]{videochatgpt}
Muhammad Maaz, Hanoona Rasheed, Salman Khan, and Fahad~Shahbaz Khan.
\newblock Video-chatgpt: Towards detailed video understanding via large vision
  and language models.
\newblock \emph{arXiv:2306.05424}, 2023.

\bibitem[Mao et~al.(2016{\natexlab{a}})Mao, Huang, Toshev, Camburu, Yuille, and
  Murphy]{okvqa}
Junhua Mao, Jonathan Huang, Alexander Toshev, Oana Camburu, Alan~L Yuille, and
  Kevin Murphy.
\newblock Generation and comprehension of unambiguous object descriptions.
\newblock In \emph{CVPR}, 2016{\natexlab{a}}.

\bibitem[Mao et~al.(2016{\natexlab{b}})Mao, Huang, Toshev, Camburu, Yuille, and
  Murphy]{refcoco}
Junhua Mao, Jonathan Huang, Alexander Toshev, Oana Camburu, Alan~L Yuille, and
  Kevin Murphy.
\newblock Generation and comprehension of unambiguous object descriptions.
\newblock In \emph{CVPR}, 2016{\natexlab{b}}.

\bibitem[Mishra et~al.(2019)Mishra, Shekhar, Singh, and Chakraborty]{ocrvqa}
Anand Mishra, Shashank Shekhar, Ajeet~Kumar Singh, and Anirban Chakraborty.
\newblock Ocr-vqa: Visual question answering by reading text in images.
\newblock In \emph{ICDAR}, 2019.

\bibitem[OpenAI(2023{\natexlab{a}})]{ChatGPT}
OpenAI.
\newblock Chatgpt.
\newblock \url{https://openai.com/blog/chatgpt/}, 2023{\natexlab{a}}.

\bibitem[OpenAI(2023{\natexlab{b}})]{GPT4}
OpenAI.
\newblock Gpt-4 technical report.
\newblock \emph{arXiv:2303.08774}, 2023{\natexlab{b}}.

\bibitem[Ouyang et~al.(2022)Ouyang, Wu, Jiang, Almeida, Wainwright, Mishkin,
  Zhang, Agarwal, Slama, Ray, et~al.]{ouyang2022training}
Long Ouyang, Jeffrey Wu, Xu Jiang, Diogo Almeida, Carroll Wainwright, Pamela
  Mishkin, Chong Zhang, Sandhini Agarwal, Katarina Slama, Alex Ray, et~al.
\newblock Training language models to follow instructions with human feedback.
\newblock In \emph{NeurIPS}, 2022.

\bibitem[Radford et~al.(2021)Radford, Kim, Hallacy, Ramesh, Goh, Agarwal,
  Sastry, Askell, Mishkin, Clark, et~al.]{CLIP}
Alec Radford, Jong~Wook Kim, Chris Hallacy, Aditya Ramesh, Gabriel Goh,
  Sandhini Agarwal, Girish Sastry, Amanda Askell, Pamela Mishkin, Jack Clark,
  et~al.
\newblock Learning transferable visual models from natural language
  supervision.
\newblock In \emph{ICML}, 2021.

\bibitem[Schwenk et~al.(2022)Schwenk, Khandelwal, Clark, Marino, and
  Mottaghi]{aokvqa}
Dustin Schwenk, Apoorv Khandelwal, Christopher Clark, Kenneth Marino, and
  Roozbeh Mottaghi.
\newblock A-okvqa: A benchmark for visual question answering using world
  knowledge.
\newblock In \emph{ECCV}, 2022.

\bibitem[Sharma et~al.(2018)Sharma, Ding, Goodman, and Soricut]{cc3m}
Piyush Sharma, Nan Ding, Sebastian Goodman, and Radu Soricut.
\newblock Conceptual captions: A cleaned, hypernymed, image alt-text dataset
  for automatic image captioning.
\newblock In \emph{ACL}, 2018.

\bibitem[Sidorov et~al.(2020)Sidorov, Hu, Rohrbach, and Singh]{textcaps}
Oleksii Sidorov, Ronghang Hu, Marcus Rohrbach, and Amanpreet Singh.
\newblock Textcaps: a dataset for image captioning with reading comprehension.
\newblock In \emph{ECCV}, 2020.

\bibitem[Singh et~al.(2019)Singh, Natarajan, Shah, Jiang, Chen, Batra, Parikh,
  and Rohrbach]{textvqa}
Amanpreet Singh, Vivek Natarajan, Meet Shah, Yu Jiang, Xinlei Chen, Dhruv
  Batra, Devi Parikh, and Marcus Rohrbach.
\newblock Towards vqa models that can read.
\newblock In \emph{CVPR}, 2019.

\bibitem[Su et~al.(2021)Su, Lu, Pan, Murtadha, Wen, and Liu]{rope}
Jianlin Su, Yu Lu, Shengfeng Pan, Ahmed Murtadha, Bo Wen, and Yunfeng Liu.
\newblock Roformer: Enhanced transformer with rotary position embedding.
\newblock \emph{arXiv:2104.09864}, 2021.

\bibitem[Taori et~al.(2023)Taori, Gulrajani, Zhang, Dubois, Li, Guestrin,
  Liang, and Hashimoto]{alpaca}
Rohan Taori, Ishaan Gulrajani, Tianyi Zhang, Yann Dubois, Xuechen Li, Carlos
  Guestrin, Percy Liang, and Tatsunori~B. Hashimoto.
\newblock Stanford alpaca: An instruction-following llama model.
\newblock \url{https://github.com/tatsu-lab/stanford_alpaca}, 2023.

\bibitem[Touvron et~al.(2023{\natexlab{a}})Touvron, Lavril, Izacard, Martinet,
  Lachaux, Lacroix, Rozi{\`e}re, Goyal, Hambro, Azhar, Rodriguez, Joulin,
  Grave, and Lample]{llama}
Hugo Touvron, Thibaut Lavril, Gautier Izacard, Xavier Martinet, Marie-Anne
  Lachaux, Timoth{\'e}e Lacroix, Baptiste Rozi{\`e}re, Naman Goyal, Eric
  Hambro, Faisal Azhar, Aurelien Rodriguez, Armand Joulin, Edouard Grave, and
  Guillaume Lample.
\newblock Llama: Open and efficient foundation language models.
\newblock \emph{arXiv:2302.13971}, 2023{\natexlab{a}}.

\bibitem[Touvron et~al.(2023{\natexlab{b}})Touvron, Martin, Stone, Albert,
  Almahairi, Babaei, Bashlykov, Batra, Bhargava, Bhosale, et~al.]{llama2}
Hugo Touvron, Louis Martin, Kevin Stone, Peter Albert, Amjad Almahairi, Yasmine
  Babaei, Nikolay Bashlykov, Soumya Batra, Prajjwal Bhargava, Shruti Bhosale,
  et~al.
\newblock Llama 2: Open foundation and fine-tuned chat models.
\newblock \emph{arXiv:2307.09288}, 2023{\natexlab{b}}.

\bibitem[Vaswani et~al.(2017)Vaswani, Shazeer, Parmar, Uszkoreit, Jones, Gomez,
  Kaiser, and Polosukhin]{vaswani2017attention}
Ashish Vaswani, Noam Shazeer, Niki Parmar, Jakob Uszkoreit, Llion Jones,
  Aidan~N Gomez, {\L}ukasz Kaiser, and Illia Polosukhin.
\newblock Attention is all you need.
\newblock In \emph{NeurIPS}, 2017.

\bibitem[Wei et~al.(2021)Wei, Bosma, Zhao, Guu, Yu, Lester, Du, Dai, and
  Le]{wei2021finetuned}
Jason Wei, Maarten Bosma, Vincent~Y Zhao, Kelvin Guu, Adams~Wei Yu, Brian
  Lester, Nan Du, Andrew~M Dai, and Quoc~V Le.
\newblock Finetuned language models are zero-shot learners.
\newblock \emph{arXiv:2109.01652}, 2021.

\bibitem[Wu et~al.(2023)Wu, Yin, Qi, Wang, Tang, and Duan]{visualchatgpt}
Chenfei Wu, Shengming Yin, Weizhen Qi, Xiaodong Wang, Zecheng Tang, and Nan
  Duan.
\newblock Visual chatgpt: Talking, drawing and editing with visual foundation
  models.
\newblock \emph{arXiv:2303.04671}, 2023.

\bibitem[Xu et~al.(2016)Xu, Mei, Yao, and Rui]{msrvtt}
Jun Xu, Tao Mei, Ting Yao, and Yong Rui.
\newblock Msr-vtt: A large video description dataset for bridging video and
  language.
\newblock In \emph{CVPR}, 2016.

\bibitem[Yang et~al.(2022)Yang, Miech, Sivic, Laptev, and Schmid]{frozenbilm}
Antoine Yang, Antoine Miech, Josef Sivic, Ivan Laptev, and Cordelia Schmid.
\newblock Zero-shot video question answering via frozen bidirectional language
  models.
\newblock In \emph{NeurIPS}, 2022.

\bibitem[Yang et~al.(2023{\natexlab{a}})Yang, Song, Li, Zhao, Ge, Li, and
  Shan]{gpt4tools}
Rui Yang, Lin Song, Yanwei Li, Sijie Zhao, Yixiao Ge, Xiu Li, and Ying Shan.
\newblock Gpt4tools: Teaching large language model to use tools via
  self-instruction.
\newblock \emph{arXiv:2305.18752}, 2023{\natexlab{a}}.

\bibitem[Yang et~al.(2023{\natexlab{b}})Yang, Li, Lin, Wang, Lin, Liu, and
  Wang]{yang2023dawn}
Zhengyuan Yang, Linjie Li, Kevin Lin, Jianfeng Wang, Chung-Ching Lin, Zicheng
  Liu, and Lijuan Wang.
\newblock The dawn of lmms: Preliminary explorations with gpt-4v (ision).
\newblock \emph{arXiv:2309.17421}, 2023{\natexlab{b}}.

\bibitem[Zhang et~al.(2023{\natexlab{a}})Zhang, Li, and Bing]{videollama}
Hang Zhang, Xin Li, and Lidong Bing.
\newblock Video-llama: An instruction-tuned audio-visual language model for
  video understanding.
\newblock \emph{arXiv:2306.02858}, 2023{\natexlab{a}}.

\bibitem[Zhang et~al.(2023{\natexlab{b}})Zhang, Han, Zhou, Hu, Yan, Lu, Li,
  Gao, and Qiao]{llamaadapter}
Renrui Zhang, Jiaming Han, Aojun Zhou, Xiangfei Hu, Shilin Yan, Pan Lu,
  Hongsheng Li, Peng Gao, and Yu Qiao.
\newblock Llama-adapter: Efficient fine-tuning of language models with
  zero-init attention.
\newblock \emph{arXiv:2303.16199}, 2023{\natexlab{b}}.

\bibitem[Zhang et~al.(2022)Zhang, Roller, Goyal, Artetxe, Chen, Chen, Dewan,
  Diab, Li, Lin, et~al.]{zhang2022opt}
Susan Zhang, Stephen Roller, Naman Goyal, Mikel Artetxe, Moya Chen, Shuohui
  Chen, Christopher Dewan, Mona Diab, Xian Li, Xi~Victoria Lin, et~al.
\newblock Opt: Open pre-trained transformer language models.
\newblock \emph{arXiv:2205.01068}, 2022.

\bibitem[Zhu et~al.(2023)Zhu, Chen, Shen, Li, and Elhoseiny]{minigpt4}
Deyao Zhu, Jun Chen, Xiaoqian Shen, Xiang Li, and Mohamed Elhoseiny.
\newblock Minigpt-4: Enhancing vision-language understanding with advanced
  large language models.
\newblock \emph{arXiv:2304.10592}, 2023.

\end{thebibliography}
}

\end{document}